\documentclass[lettersize,journal]{IEEEtran}
\usepackage[T1]{fontenc}
\usepackage{amsmath,amsfonts}
\usepackage{amssymb}
\usepackage{amsthm}
\usepackage[ruled,linesnumbered]{algorithm2e}
\usepackage{array}
\usepackage[caption=false,font=normalsize,labelfont=sf,textfont=sf]{subfig}
\usepackage{textcomp}
\usepackage{stfloats}
\usepackage{url}
\usepackage{verbatim}
\usepackage{graphicx}
\usepackage{multirow}
\usepackage{booktabs}
\usepackage{adjustbox}
\usepackage{diagbox}
\usepackage{tabularx}
\usepackage[table]{xcolor}
\usepackage{supertabular}
\usepackage{cite}
\usepackage{siunitx}
\usepackage{exscale}
\usepackage{relsize}
\usepackage{ltxcmds}
\usepackage{subscript}
\usepackage{bm}
\usepackage{nomencl}
\usepackage{threeparttable}
\usepackage{balance}
\graphicspath{{./figs/}}
\makenomenclature

\makeatletter
\newcommand*\bigcdot{\mathpalette\bigcdot@{.5}}
\newcommand*\bigcdot@[2]{\mathbin{\vcenter{\hbox{\scalebox{#2}{$\m@th#1\bullet$}}}}}
\makeatother
\newtheorem{lemma}{Lemma}     
\newtheorem{theorem}{Theorem}
\newtheorem{corollary}{Corollary}

\theoremstyle{definition}

\newtheorem{definition}{Definition}
\newtheorem{assumption}{Assumption}
\newenvironment{pf}{{\noindent\it Proof}\quad}{\hfill $\blacksquare$\par}% Keep lemma linked to theorem counter
\newtheorem{rmk}{Remark} 

\newtheoremstyle{myTheo_style1}%
  {3pt}%
  {0pt}%
  {}%
  {}%
  {\itshape}%
  {}%
  {5pt}%
  {}%

\newtheoremstyle{myTheo_style2}%
  {3pt}%
  {0pt}%
  {}%
  {}%
  {\itshape}%
  {:}%
  {5pt}%
  {}%

\newtheoremstyle{myProof_style}%
  {0pt}%
  {0pt}%
  {}%
  {}%
  {\itshape}%
  {:}%
  {5pt}%
  {}%

\newtheoremstyle{myAssumption_style}%
  {0pt}%
  {0pt}%
  {}%
  {}%
  {\itshape}%
  {:}%
  {5pt}%
  {}%
\theoremstyle{myTheo_style1}

\theoremstyle{myTheo_style2}

\theoremstyle{myTheo_style1}
\theoremstyle{myProof_style}

\theoremstyle{myAssumption_style}

\theoremstyle{myProof_style}

\theoremstyle{myProof_style}

\def\BibTeX{{\rm B\kern-.05em{\sc i\kern-.025em b}\kern-.08em
    T\kern-.1667em\lower.7ex\hbox{E}\kern-.125emX}}

\begin{document}

%瑕佷笉瑕佸啓瀹夊叏鎬ц瘎浼板拰瀹夊叏鎺у埗
% \title{SafeLink: Data-Driven Safety-Critical Control for Irregular Unsafe Regions With sudden Changes}
\title{Learning-based Adaptive Safety-Critical Control With Evolving Unsafe Regions}

\author{
Songqiao Hu,
Zidong Wang,~\IEEEmembership{Fellow,~IEEE},
Zeyi Liu,
Zhen Shen,
and Xiao He,~\IEEEmembership{Senior Member,~IEEE}
\thanks{
This work was supported in part by the National Natural Science Foundation of China under Grants 62525308, 62473223, 52172323, and 624B2087, in part by the Beijing Natural Science Foundation under Grant L241016. (Corresponding author: Xiao He.)
}
\thanks{
Songqiao Hu, Zeyi Liu, and Xiao He are with the Department of Automation and the Institute for Embodied Intelligence and Robotics, Tsinghua University, Beijing 100084, China (e-mails: hsq23@mails.tsinghua.edu.cn; liuzy21@mails.tsinghua.edu.cn; hexiao@tsinghua.edu.cn).
}
\thanks{
Zidong Wang is with the Department of Computer Science, Brunel University London, UB8 3PH Uxbridge, U.K. (e-mail: zidong.wang@brunel.ac.uk).
}
\thanks{
Zhen Shen is with the State Key Laboratory of Multimodal Artificial Intelligence Systems, Beijing Engineering Research Center of Intelligent Systems and Technology, Institute of Automation, Chinese Academy of Sciences, Beijing, China (e-mail: zhen.shen@ia.ac.cn).
}
}

\maketitle

\begin{abstract}
Control barrier functions (CBFs) provide a principled framework for
safety-critical control, but their construction typically requires an
explicit and differentiable description of the safe or unsafe region.
It becomes challenging for data-defined unsafe regions that may evolve over time. This paper proposes SafeLink, a data-driven CBF
construction and adaptation method based on a cost-sensitive random
vector functional link (RVFL) network. SafeLink introduces asymmetric
misclassification costs to promote conservative unsafe-region
representation while preserving a closed-form solution. We establish
the Lipschitz continuity of the learned CBF and its derivatives, and
derive sufficient conditions for conservative unsafe-region coverage
and the corresponding interval-wise safety guarantees. Analytical updates are
further developed for adjusting the misclassification cost and for
incrementally adding or decrementally removing samples, avoiding full
retraining when the unsafe region changes. Experiments on a two-link
manipulator demonstrate that SafeLink rapidly adapts to evolving unsafe regions, enables collision-free target reaching, and
achieves substantially lower update runtimes than baselines.
\end{abstract}

\begin{IEEEkeywords}
Control barrier function, safety-critical control, incremental learning,
random vector functional link network, robotic manipulators.
\end{IEEEkeywords}
\section{Introduction}

\IEEEPARstart{T}{he} safety of dynamic systems has long been a central concern in autonomous systems and is generally associated with preventing harm to personnel, equipment, or the environment \cite{liu2023real}. As autonomous systems are increasingly deployed in complex and safety-critical environments, ensuring safe operation has become essential for the reliable execution of tasks by intelligent agents \cite{liu2025safety,alan2023control,11007216}. This has motivated extensive research on safety-critical control methods that explicitly incorporate safety requirements into control synthesis \cite{lacevic2013safety,10510639}.

Safety requirements for control systems are commonly formulated as constraints on the system state and represented through forward-invariant sets \cite{blanchini1999set,clark2020control,10976622}. Within this framework, CBFs have emerged as a promising approach for safety enforcement owing to their formal safety guarantees and real-time implementability \cite{ames2016control,ames2019control,dawson2023safe}. CBFs encode safety requirements as constraints involving the system dynamics and control inputs, thereby enabling safe control actions to be synthesized online \cite{tee2009barrier,prajna2004safety,wu2025safety}. Given a predefined safe set and a nominal controller that does not explicitly account for safety, CBF-based methods commonly formulate a quadratic programming (QP) problem, in which the CBF condition is imposed as a constraint while the deviation from the nominal control input is minimized. Building on this paradigm, CBFs have been successfully applied to robotic manipulation \cite{wang2025constrained}, autonomous driving \cite{xiao2023barriernet}, unmanned aerial vehicle control \cite{kang2025control}, and satellite trajectory optimization \cite{breeden2023robust}, demonstrating their broad applicability.

Conventional model-based CBF construction typically relies on an explicit and sufficiently differentiable representation of the safe or unsafe region \cite{huang2025dynamic,rauscher2016constrained,breeden2023robust,sun2024safety,xie2025cbf,11017678}. Once such a representation is available, safety constraints can be incorporated into control synthesis through the corresponding CBF conditions. However, owing to the complexity and evolving nature of operating environments, unsafe regions may exhibit irregular geometries and continuously evolve, making explicit analytical descriptions difficult to obtain and maintain. Even when analytical expressions can be constructed, irregular boundaries may lead to nonsmooth or piecewise-defined representations, complicating their use in conventional CBF and high-order CBF (HOCBF) formulations that require sufficient differentiability. Moreover, in many practical scenarios, safety information is available only through discrete observations or sampled data, such as obstacle measurements from LiDAR or 3D point clouds from depth cameras. These challenges have motivated increasing interest in data-driven approaches for learning safe or unsafe regions and constructing corresponding CBFs \cite{cheng2019end,guerrier2024learning,dawson2022safe}.

Motivated by these challenges, recent studies have explored data-driven CBF construction from labeled state samples using machine learning techniques. Safety labels can be obtained from sensor measurements, expert knowledge, or human-in-the-loop feedback, after which classifiers such as support vector machines (SVMs) and multilayer perceptrons (MLPs) can be employed to approximate safe and unsafe regions and construct corresponding CBFs \cite{srinivasan2020synthesis,xiao2023learning,robey2020learning,chen2022machine,zhang2023exact}. Despite these advances, several challenges remain. First, conventional classification objectives generally do not explicitly distinguish the safety consequences of different misclassification errors, whereas classifying an unsafe state as safe is considerably more critical than the converse in safety-critical control. Second, when safety constraints or hazard regions change suddenly \cite{liu2024online,11104131}, batch retraining can incur substantial computational overhead and hinder rapid online adaptation. Third, although nonlinear learning models can provide flexible function approximation, their training and parameter updates generally lack a closed-form structure, making exact online adaptation and analytical safety analysis less tractable. These limitations motivate a CBF learning framework that combines safety-aware asymmetric classification, analytical tractability, and efficient adaptation to evolving unsafe regions.

To address these challenges, we propose SafeLink, a data-driven safety-critical control framework for constructing and adapting CBFs for evolving unsafe regions. SafeLink is built on the RVFL network, whose fixed random feature mapping and closed-form output solution provide both nonlinear approximation capability and analytical tractability \cite{pao1994learning,igelnik1995stochastic,needell2024random}. The main contributions of this work are as follows:
\begin{enumerate}[]
\item A cost-sensitive RVFL-based CBF construction method is developed for complex data-defined unsafe regions. By introducing asymmetric misclassification costs, the proposed formulation accounts for the different safety consequences of classification errors while preserving a closed-form solution.

\item The learned CBF is theoretically analyzed by establishing the Lipschitz continuity of the CBF and its derivatives, and deriving sufficient conditions for conservative unsafe-region coverage and interval-wise safety guarantees
under evolving unsafe regions.

\item Analytical adaptation rules are derived for adjusting the misclassification cost and for incrementally adding or decrementally removing samples. These rules avoid  retraining from scratch and enable rapid adaptation of the learned CBF to evolving unsafe regions.
\end{enumerate}

The remainder of this paper reviews the preliminaries in Section~\ref{sec:pre}, presents SafeLink in Section~\ref{sec:methods}, reports experiments in Section~\ref{sec:experiments}, and concludes in Section~\ref{sec:con}.

\section{Preliminaries}
\label{sec:pre}
\subsection{Control Barrier Functions}

Consider the following control-affine system:
\begin{equation}
\label{eq:system}
\dot{\bm{x}}
=
\bm{f}(\bm{x})
+
\bm{g}(\bm{x})\bm{u},
\end{equation}
where $\bm{x}\in\mathbb{R}^n$ is the system state,
$\bm{f}:\mathbb{R}^n\rightarrow\mathbb{R}^n$ and
$\bm{g}:\mathbb{R}^n\rightarrow\mathbb{R}^{n\times q}$
are locally Lipschitz continuous vector fields, and
$\bm{u}\in U\subseteq\mathbb{R}^q$ is the control input.

The admissible input set is
\begin{equation}
\label{eq:control_bound}
U
:=
\left\{
\bm{u}\in\mathbb{R}^q:
\bm{u}_{\min}\leq\bm{u}\leq\bm{u}_{\max}
\right\},
\end{equation}
where the inequalities are interpreted element-wise.

\begin{definition}[\textit{Unsafe region}]
The real unsafe region $\mathcal{R}_u\subseteq\mathbb{R}^n$ is the set of states that violate the safety requirements or physical constraints of the system.
A state $\bm{x}$ is unsafe if $\bm{x}\in\mathcal{R}_u$.
\end{definition}

\begin{definition}[\textit{Relative degree}]
The relative degree of a differentiable function
$B:\mathbb{R}^n\rightarrow\mathbb{R}$ with respect to
system~\eqref{eq:system} is the number of times that $B$
must be differentiated along the system dynamics before the
control input $\bm{u}$ explicitly appears.
\end{definition}

\begin{definition}[\textit{Forward invariance}]
A set $\mathcal{C}\subseteq\mathbb{R}^n$ is forward
invariant for system~\eqref{eq:system} if every solution
starting from $\bm{x}(t_0)\in\mathcal{C}$ satisfies
$\bm{x}(t)\in\mathcal{C}$ for all $t\geq t_0$.
\end{definition}

Let $B:\mathbb{R}^n\rightarrow\mathbb{R}$ be continuously
differentiable. The barrier-induced safe and unsafe sets are
defined as
\begin{equation}
\label{eq:invariant_set}
\begin{aligned}
\mathcal{C}_1
&:=
\left\{
\bm{x}\in\mathbb{R}^n:
B(\bm{x})\geq0
\right\},\\
\widehat{\mathcal{R}}_u
&:=
\mathbb{R}^n\setminus\mathcal{C}_1
=
\left\{
\bm{x}\in\mathbb{R}^n:
B(\bm{x})<0
\right\}.
\end{aligned}
\end{equation}

Specifically, $\mathcal{R}_u$ denotes the real unsafe region,
whereas $\widehat{\mathcal{R}}_u$ is the unsafe region
induced by $B$. Sufficient conditions for the conservative inclusion $\mathcal{R}_u\subseteq\widehat{\mathcal{R}}_u$ will be established later.

\begin{definition}[\textit{Control barrier function}
\cite{ames2019control}]
\label{def:CBF}
Given the safe set $\mathcal{C}_1$ in
Eq.~\eqref{eq:invariant_set}, the function $B$ is a control
barrier function for system~\eqref{eq:system} if there
exists a class-$\mathcal{K}$ function $\alpha$ such that
\begin{equation}
\label{eq:constraints}
\sup_{\bm{u}\in U}
\left[
L_fB(\bm{x})
+
L_gB(\bm{x})\bm{u}
+
\alpha\bigl(B(\bm{x})\bigr)
\right]
\geq0,
\quad
\forall\bm{x}\in\mathcal{C}_1.
\end{equation}
\end{definition}

\begin{lemma}[\!\!\!\cite{ames2014control}]
\label{lem:CBF_invariance}
Let $B$ be a CBF as defined in
Def.~\ref{def:CBF}. If
$\bm{x}(t_0)\in\mathcal{C}_1$, then any
Lipschitz-continuous control input $\bm{u}(t)\in U$
satisfying the constraint in Eq.~\eqref{eq:constraints} for
all $t\geq t_0$ renders $\mathcal{C}_1$ forward invariant
for system~\eqref{eq:system}.
\end{lemma}

\subsection{High-Order Control Barrier Functions}

Consider a sufficiently differentiable function
$B:\mathbb{R}^n\rightarrow\mathbb{R}$ with relative degree
$r$ for system~\eqref{eq:system}. Define
$\psi_0(\bm{x}):=B(\bm{x})$ and recursively construct
\begin{equation}
\label{eq:psi}
\psi_i(\bm{x})
:=
\dot{\psi}_{i-1}(\bm{x})
+
\alpha_i\bigl(\psi_{i-1}(\bm{x})\bigr),
\quad i\in\{1,\ldots,r-1\},
\end{equation}
where
$\dot{\psi}_{i-1}(\bm{x})
=
\frac{\partial\psi_{i-1}}{\partial\bm{x}}
\dot{\bm{x}}$
and each $\alpha_i$ is a class-$\mathcal K$ function.

The associated sets are defined as
\begin{equation}
\label{eq:psi_C}
\mathcal{C}_i
:=
\left\{
\bm{x}\in\mathbb{R}^n:
\psi_{i-1}(\bm{x})\geq0
\right\},
\quad i\in\{1,\ldots,r\}.
\end{equation}

Since $\psi_0(\bm{x})=B(\bm{x})$, the first HOCBF set is
\(
\mathcal{C}_1
=
\left\{
\bm{x}\in\mathbb{R}^n:
B(\bm{x})\geq0
\right\},
\)
and the corresponding barrier-induced unsafe set is
$\widehat{\mathcal{R}}_u
:=
\mathbb{R}^n\setminus\mathcal{C}_1
=
\{\bm{x}\in\mathbb{R}^n:B(\bm{x})<0\}$.

\begin{definition}[\textit{High-order control barrier function} \cite{xiao2021high}]
\label{def:HOCBF}
Let $\mathcal{C}_1,\ldots,\mathcal{C}_r$ and
$\psi_1,\ldots,\psi_{r-1}$ be defined by
Eqs.~\eqref{eq:psi_C} and~\eqref{eq:psi}, respectively.
The function $B$ is a high-order control barrier function
 of relative degree $r$ for system~\eqref{eq:system}
if there exist differentiable class-$\mathcal K$ functions
$\alpha_i$, $i\in\{1,\ldots,r-1\}$, and a class-$\mathcal K$
function $\alpha_r$ such that
\begin{equation}
\label{eq:u_HOCBF}
\begin{aligned}
\sup_{\bm{u}\in U}\Big[
&L_f^rB(\bm{x})
+L_gL_f^{r-1}B(\bm{x})\bm{u}
+\mathcal{O}(B(\bm{x})) \\
&+\alpha_r\bigl(\psi_{r-1}(\bm{x})\bigr)
\Big]
\geq0,
\quad
\forall\bm{x}\in\bigcap_{i=1}^{r}\mathcal{C}_i,
\end{aligned}
\end{equation}
where
$\mathcal{O}(B(\bm{x}))
=
\sum_{i=1}^{r-1}
L_f^i
(\alpha_{r-i}\circ\psi_{r-i-1})(\bm{x})$.
Moreover,
$L_gL_f^{r-1}B(\bm{x})\neq0$
on the boundary of
$\bigcap_{i=1}^{r}\mathcal{C}_i$.
\end{definition}

\begin{lemma}[\!\!\cite{xiao2021high}]
Given a HOCBF $B$ as defined in
Def.~\ref{def:HOCBF}, suppose that
$\bm{x}(t_0)\in\bigcap_{i=1}^{r}\mathcal{C}_i$.
Any Lipschitz-continuous control input
$\bm{u}(t)\in U$ satisfying
\begin{equation}
\label{eq:u_constraint}
\begin{aligned}
L_f^rB(\bm{x})
&+L_gL_f^{r-1}B(\bm{x})\bm{u}(t)\\
&+\mathcal{O}(B(\bm{x}))
+\alpha_r\bigl(\psi_{r-1}(\bm{x})\bigr)
\geq0
\end{aligned}
\end{equation}
for all $t\geq t_0$ renders
$\bigcap_{i=1}^{r}\mathcal{C}_i$
forward invariant for system~\eqref{eq:system}.
\end{lemma}

Let $\bm{u}_r$ denote a reference input designed for the
control objective without explicitly accounting for safety.
The safety-filtered input is obtained by solving
\begin{equation}
\label{eq:solve}
\begin{aligned}
\bm{u}_{\mathrm{safe}}
&=
\operatorname*{arg\,min}_{\bm{u}\in U}
\|\bm{u}-\bm{u}_r\|_2^2
\\
\mathrm{s.t.}\quad
&
L_f^r B(\bm{x})
+
L_gL_f^{r-1}B(\bm{x})\bm{u}
\\
&
\quad+
\mathcal{O}(B(\bm{x}))
+
\alpha_r\bigl(\psi_{r-1}(\bm{x})\bigr)
\geq 0 .
\end{aligned}
\end{equation}

For $r=1$, $\mathcal{O}(B(\bm{x}))=0$ and
$\psi_{r-1}(\bm{x})=B(\bm{x})$, so
Eq.~\eqref{eq:solve} reduces to the standard CBF-QP.
\subsection{Random Vector Functional Link Network}

The RVFL network is a shallow
neural network consisting of an input layer, a randomly
initialized enhancement layer, and a trainable linear output
layer~\cite{pao1994learning}. Let $M$ denote the number of
enhancement nodes. For any state
$\bm{x}\in\mathbb{R}^n$, define the extended feature vector as
\begin{equation}
\label{eq:extend}
\widetilde{\bm{x}}
:=
\begin{bmatrix}
\bm{x}\\
\phi\!\left(\bm{W}_e^\top\bm{x}+\bm{b}_e\right)
\end{bmatrix}
\in\mathbb{R}^{n+M},
\end{equation}
where $\bm{W}_e\in\mathbb{R}^{n\times M}$ and
$\bm{b}_e\in\mathbb{R}^{M}$ are randomly initialized and fixed,
and $\phi(\cdot)$ is applied element-wise.

For each labeled sample, define the one-hot safety label by
\begin{equation}
\label{eq:safety_label}
\widetilde{\bm{y}}
:=
\begin{cases}
[1,\ 0]^\top, & \bm{x}\in\mathcal{R}_u,\\
[0,\ 1]^\top, & \bm{x}\in\mathbb{R}^n\setminus\mathcal{R}_u.
\end{cases}
\end{equation}

Given $N$ labeled samples
$\{(\bm{x}_i,\widetilde{\bm{y}}_i)\}_{i=1}^{N}$, define the
data matrix $\bm{A}$ and the label matrix $\bm{Y}$ by
\begin{equation}
\label{eq:A_Y}
\bm{A}
:=
\begin{bmatrix}
\widetilde{\bm{x}}_1^\top\\
\vdots\\
\widetilde{\bm{x}}_N^\top
\end{bmatrix}
\in\mathbb{R}^{N\times(n+M)},
\qquad
\bm{Y}
:=
\begin{bmatrix}
\widetilde{\bm{y}}_1^\top\\
\vdots\\
\widetilde{\bm{y}}_N^\top
\end{bmatrix}
\in\mathbb{R}^{N\times2}.
\end{equation}

The output weight matrix
$\bm{W}_b\in\mathbb{R}^{(n+M)\times2}$ is obtained by ridge
regression,
\begin{equation}
\label{eq:optimization}
\bm{W}_b
=
\operatorname*{arg\,min}_{\bm{W}}
\left\{
\lambda\|\bm{W}\|_F^2
+
\|\bm{A}\bm{W}-\bm{Y}\|_F^2
\right\},
\end{equation}
where $\lambda>0$ is the regularization parameter. The
closed-form solution is
\begin{equation}
\label{eq:W_b}
\bm{W}_b
=
\left(
\lambda\bm{I}
+
\bm{A}^\top\bm{A}
\right)^{-1}
\bm{A}^\top\bm{Y}.
\end{equation}

For a state $\bm{x}$, the RVFL prediction is
\begin{equation}
\label{eq:rvfl_prediction}
\widehat{\bm{Y}}(\bm{x})
=
\widetilde{\bm{x}}^\top\bm{W}_b
=
\begin{bmatrix}
\widehat{Y}_1(\bm{x}) & \widehat{Y}_2(\bm{x})
\end{bmatrix}.
\end{equation}
The state is classified as unsafe if
$\widehat{Y}_1(\bm{x})>\widehat{Y}_2(\bm{x})$, and as safe
otherwise.

\begin{figure*}[htbp]
    \centering
    \begin{center}
\includegraphics[width=\linewidth]{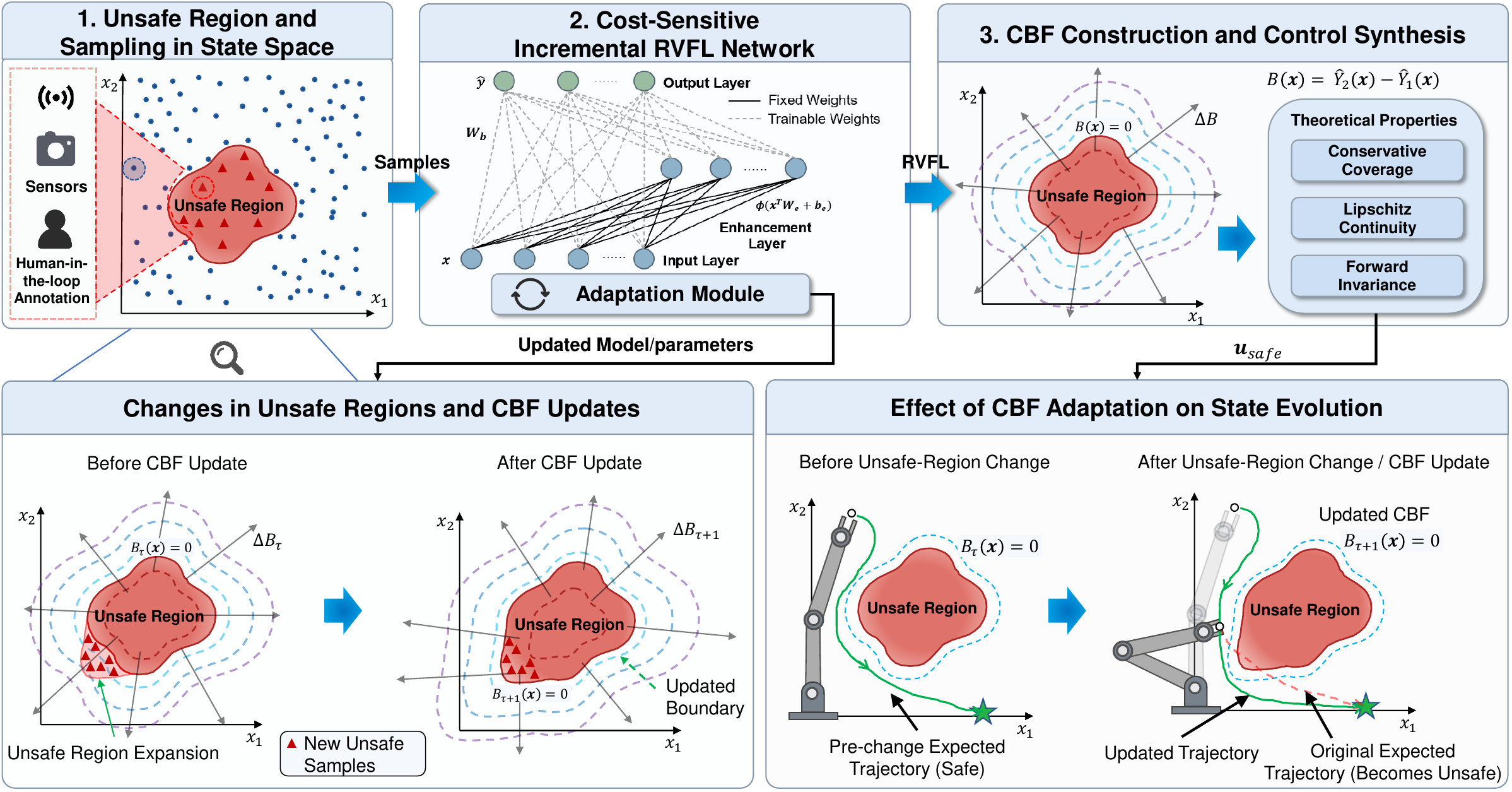}  
    \end{center}
\caption{Overview of the proposed SafeLink framework. SafeLink constructs a differentiable CBF from labeled safety samples using a cost-sensitive RVFL model, and performs exact incremental/decremental updates when unsafe regions evolve.}
\label{fig:diagram}
\end{figure*}

\section{The Proposed Framework}
This section presents the SafeLink framework. We first construct a differentiable CBF using a cost-sensitive RVFL model, then analyze its safety properties, and finally derive analytical updates for adaptation to evolving unsafe regions. An overview of the framework is illustrated in Fig.~\ref{fig:diagram}.
\label{sec:methods}

\subsection{SafeLink Construction for Static Unsafe Region}
\label{subsec:safelink_construction}

For a static unsafe region $\mathcal{R}_u$, SafeLink constructs
a differentiable barrier function from labeled state samples
through a cost-sensitive RVFL model. Let $c_1$ and $c_2$
denote the costs of misclassifying an unsafe state as safe
and a safe state as unsafe, respectively. The cost matrix is
defined as
\begin{equation}
\label{eq:cost_matrix}
\bm{C}
:=
\begin{bmatrix}
0 & c_1\\
c_2 & 0
\end{bmatrix},
\end{equation}
where $c_1>c_2$ is typically selected in safety-critical applications to penalize misclassifying unsafe states as safe more heavily.

To incorporate the asymmetric misclassification costs while
preserving the closed-form RVFL training structure, the
ridge-regression objective in Eq.~\eqref{eq:optimization} is
augmented as
\begin{equation}
\label{eq:cost_function}
\begin{aligned}
\bm{W}_b
=
\operatorname*{arg\,min}_{\bm{W}}
\Big\{
&\lambda\|\bm{W}\|_F^2
+\|\bm{A}\bm{W}-\bm{Y}\|_F^2\\
&+2\operatorname{tr}
\left(
\bm{A}\bm{W}\bm{C}^{\top}\bm{Y}^{\top}
\right)
\Big\}.
\end{aligned}
\end{equation}

The trace term introduces the prescribed asymmetric
classification costs while retaining the quadratic structure
of the optimization problem. For example, an idealized false-safe prediction $[0,1]^\top$ for an unsafe sample $[1,0]^\top$ is associated with the cost $c_1$, whereas the opposite misclassification is associated with $c_2$.

Since $\lambda>0$, Eq.~\eqref{eq:cost_function} is strictly
convex and admits the unique closed-form solution
\begin{equation}
\label{eq:W_b_we}
\bm{W}_b
=
\underbrace{
\left(
\lambda\bm{I}+\bm{A}^{\top}\bm{A}
\right)^{-1}
}_{\bm{K}}
\underbrace{
\bm{A}^{\top}\bm{Y}(\bm{I}-\bm{C})
}_{\bm{Q}}
=
\bm{K}\bm{Q}.
\end{equation}

For a state $\bm{x}$, the corresponding RVFL output is
\begin{equation}
\label{eq:Y_hat}
\widehat{\bm{Y}}(\bm{x})
=
\widetilde{\bm{x}}^{\top}\bm{W}_b
=
\begin{bmatrix}
\widehat{Y}_1(\bm{x}) &
\widehat{Y}_2(\bm{x})
\end{bmatrix}.
\end{equation}

According to the labeling convention in
Eq.~\eqref{eq:safety_label},
$\widehat{Y}_1(\bm{x})$ and $\widehat{Y}_2(\bm{x})$
represent the unsafe and safe prediction scores,
respectively. Define
\begin{equation}
\label{eq:v_B}
\bm{v}_B
:=
\begin{bmatrix}
-1 & 1
\end{bmatrix}^{\top}.
\end{equation}
The scalar barrier function is constructed as
\begin{equation}
\label{eq:CBF}
B(\bm{x})=\widehat{\bm{Y}}(\bm{x})\bm{v}_B=
\widehat{Y}_2(\bm{x})
-
\widehat{Y}_1(\bm{x}).
\end{equation}

Therefore,
$B(\bm{x})<0$ if and only if
$\widehat{Y}_1(\bm{x})>\widehat{Y}_2(\bm{x})$,
which is consistent with the unsafe-state classification rule
of the RVFL model. Accordingly, $B(\bm{x})\geq0$ defines the
barrier-induced safe set $\mathcal{C}_1$ in
Eq.~\eqref{eq:invariant_set}.

For subsequent analysis and control implementation, let
$\bm{w}_{e,k}\in\mathbb{R}^n$ denote the random input weight
of the $k$-th enhancement node and $b_{e,k}$ its bias.
Let $\bm{W}_{b_0}\in\mathbb{R}^{n\times2}$ contain the rows
of $\bm{W}_b$ associated with the direct input features, and
let $\bm{w}_{b,k}\in\mathbb{R}^{1\times2}$ denote the output
weights associated with the $k$-th enhancement node. Define
\begin{equation}
\label{eq:compact_definitions}
\begin{aligned}
s_k(\bm{x})
&:=
\bm{x}^{\top}\bm{w}_{e,k}+b_{e,k},\\
\bm{\beta}_0
&:=
\bm{W}_{b_0}\bm{v}_B,\\
\beta_k
&:=
\bm{w}_{b,k}\bm{v}_B.
\end{aligned}
\end{equation}
Then the learned barrier can be expressed as
\begin{equation}
\label{eq:barrier_scalar_form}
B(\bm{x})
=
\bm{x}^{\top}\bm{\beta}_0
+
\sum_{k=1}^{M}
\beta_k
\phi\!\left(s_k(\bm{x})\right).
\end{equation}

Its first- and second-order derivatives are
\begin{equation}
\label{eq:barrier_gradient}
\nabla_{\bm{x}}B(\bm{x})
=
\bm{\beta}_0
+
\sum_{k=1}^{M}
\beta_k
\phi'\!\left(s_k(\bm{x})\right)
\bm{w}_{e,k},
\end{equation}
and
\begin{equation}
\label{eq:barrier_hessian}
\nabla_{\bm{x}}^2B(\bm{x})
=
\sum_{k=1}^{M}
\beta_k
\phi''\!\left(s_k(\bm{x})\right)
\bm{w}_{e,k}\bm{w}_{e,k}^{\top}.
\end{equation}

For a barrier of relative degree $r$, the required Lie derivatives are obtained from the learned CBF and incorporated into the HOCBF-QP in Eq.~\eqref{eq:solve}. Accordingly, the activation function is selected to be at least $r$ times differentiable.

The following result establishes the Lipschitz continuity of the learned barrier and its derivatives.

\begin{theorem}
\label{thm:continues}
Suppose that
$\phi(\cdot)$ is $r$ times differentiable and
$\phi^{(p)}(\cdot)$ is Lipschitz continuous for every
$p\in\{0,1,\ldots,r\}$, where
$\phi^{(0)}(\cdot):=\phi(\cdot)$. Then $B(\bm{x})$ and its derivatives
$\nabla_{\bm{x}}^pB(\bm{x})$,
$p\in\{1,\ldots,r\}$, are Lipschitz continuous.
\end{theorem}

\begin{pf}
Let $L_{\phi}$ denote a Lipschitz constant of $\phi$. For any $\bm{x}_1,\bm{x}_2\in\mathbb{R}^n$, it follows from Eq.~\eqref{eq:barrier_scalar_form} that
\begin{align}
|B(\bm{x}_1)-B(\bm{x}_2)|
&\leq
\|\bm{\beta}_0\|_2
\|\bm{x}_1-\bm{x}_2\|_2
\nonumber\\
&\quad+
\sum_{k=1}^{M}
|\beta_k|L_{\phi}
\left|
(\bm{x}_1-\bm{x}_2)^\top\bm{w}_{e,k}
\right|
\nonumber\\
&\leq
L_B\|\bm{x}_1-\bm{x}_2\|_2,
\label{eq:B_lipschitz_bound}
\end{align}
where
\begin{equation}
\label{eq:B_lipschitz_constant}
L_B
=
\|\bm{\beta}_0\|_2
+
\sum_{k=1}^{M}
L_{\phi}|\beta_k|
\|\bm{w}_{e,k}\|_2.
\end{equation}
Thus, $B(\bm{x})$ is Lipschitz continuous.

For any $p\in\{1,\ldots,r\}$, differentiating Eq.~\eqref{eq:barrier_scalar_form} gives
\begin{equation}
\label{eq:barrier_pth_derivative}
\nabla_{\bm{x}}^pB(\bm{x})
=
\bm{\Gamma}_p
+
\sum_{k=1}^{M}
\beta_k
\phi^{(p)}\!\left(s_k(\bm{x})\right)
\bm{w}_{e,k}^{\otimes p},
\end{equation}
where $\bm{\Gamma}_1=\bm{\beta}_0$, $\bm{\Gamma}_p=0$ for $p\geq2$, and $\bm{w}_{e,k}^{\otimes p}$ denotes the $p$-fold tensor product. For higher-order derivative tensors, $\|\cdot\|_F$ denotes the Frobenius norm.

Let $L_{\phi^{(p)}}$ denote a Lipschitz constant of $\phi^{(p)}$. Since $\bm{\Gamma}_p$ is independent of $\bm{x}$ and $\|\bm{w}_{e,k}^{\otimes p}\|_F=\|\bm{w}_{e,k}\|_2^p$, for any $\bm{x}_1,\bm{x}_2\in\mathbb{R}^n$,
\begin{align}
&
\left\|
\nabla_{\bm{x}}^pB(\bm{x}_1)
-
\nabla_{\bm{x}}^pB(\bm{x}_2)
\right\|_F
\nonumber\\
&\leq
\sum_{k=1}^{M}
|\beta_k|L_{\phi^{(p)}}
\left|
(\bm{x}_1-\bm{x}_2)^\top\bm{w}_{e,k}
\right|
\|\bm{w}_{e,k}\|_2^p
\nonumber\\
&\leq
L_{\nabla_{\bm{x}}^pB}
\|\bm{x}_1-\bm{x}_2\|_2,
\label{eq:derivative_lipschitz_bound}
\end{align}
where
\begin{equation}
\label{eq:derivative_lipschitz_constant}
L_{\nabla_{\bm{x}}^pB}
=
\sum_{k=1}^{M}
L_{\phi^{(p)}}|\beta_k|
\|\bm{w}_{e,k}\|_2^{p+1}.
\end{equation}
Hence, $\nabla_{\bm{x}}^pB(\bm{x})$ is Lipschitz continuous for every $p\in\{1,\ldots,r\}$. Together with the Lipschitz continuity of $B(\bm{x})$, this proves the result.
\end{pf}

\color{black}

\subsection{Safety Analysis of SafeLink}
\label{subsec:safety_analysis}

We next analyze the safety properties of SafeLink. We first establish conditions under which the learned unsafe set conservatively covers the real unsafe region, and then derive an interval-wise safety guarantee.

Let $\mathcal{X}_u$ and $\mathcal{X}_s$ denote the current sets of
unsafe and safe training samples, respectively, and let
$\bm{x}_c$ denote the current state. If
$\bm{x}_c\notin\mathcal{R}_u$, it is appended to the training
set with the safe label $[0,1]^\top$. The augmented safe sample
set is
\begin{equation}
\label{eq:augmented_safe_set}
\mathcal{X}_s^+
:=
\mathcal{X}_s\cup\{\bm{x}_c\}.
\end{equation}

Accordingly, let $\bm{A}$ denote the extended feature matrix
constructed from the augmented training set
$\mathcal{X}_u\cup\mathcal{X}_s^+$. Define
\begin{equation}
\label{eq:hat_matrix}
\bm{H}
:=
\bm{A}
\left(
\lambda\bm{I}
+
\bm{A}^{\top}\bm{A}
\right)^{-1}
\bm{A}^{\top},
\end{equation}
where $h_{ki}$ denotes the $(k,i)$-th element of $\bm{H}$. 

For each sample $\bm{x}_k$, define
\begin{equation}
\label{eq:aggregate_influences}
S_k
:=
\sum_{\bm{x}_i\in\mathcal{X}_u} h_{ki},
\qquad
T_k
:=
\sum_{\bm{x}_j\in\mathcal{X}_s^+} h_{kj}.
\end{equation}

Here, $S_k$ and $T_k$ collect the contributions of the unsafe
and safe samples, respectively, to the prediction at
$\bm{x}_k$. In particular, $S_c$ and $T_c$ denote the
corresponding quantities at $\bm{x}_c$.

\begin{assumption}
\label{ass:1}
The current state satisfies
$\bm{x}_c\notin\mathcal{R}_u$, and
$S_k>0$ for every $\bm{x}_k\in\mathcal{X}_u$.
Given prescribed CBF margins
$\varepsilon_u>0$ and $\varepsilon_s\geq0$, define
\begin{equation}
\label{eq:unsafe_cost_bound}
\underline{c}_1
:=
\max_{\bm{x}_k\in\mathcal{X}_u}
\left\{
\frac{(1+c_2)T_k+\varepsilon_u}{S_k}-1
\right\},
\end{equation}
and
\begin{equation}
\label{eq:current_cost_set}
\mathcal{I}_c
:=
\left\{
c_1\geq0:
-(1+c_1)S_c+(1+c_2)T_c
\geq\varepsilon_s
\right\}.
\end{equation}
The following compatibility condition holds:
\begin{equation}
\label{eq:cost_compatibility}
\left[
\max\{0,\underline{c}_1\},
\infty
\right)
\cap
\mathcal{I}_c
\neq\emptyset.
\end{equation}
\end{assumption}

Assumption~\ref{ass:1} can be directly checked from the
augmented training data. The bound $\underline{c}_1$ enforces
the prescribed negative margin at all unsafe training samples,
whereas $\mathcal{I}_c$ ensures the nonnegative margin at the
current state. For $S_c\neq0$, define
\begin{equation}
\label{eq:current_cost_threshold}
c_{1,c}
:=
\frac{(1+c_2)T_c-\varepsilon_s}{S_c}-1.
\end{equation}

Then $c_1\leq c_{1,c}$ is required when $S_c>0$, whereas
$c_1\geq c_{1,c}$ is required when $S_c<0$. If $S_c=0$,
the condition reduces to $(1+c_2)T_c\geq\varepsilon_s$.
Thus, Eq.~\eqref{eq:cost_compatibility} characterizes the
existence of a compatible misclassification cost.

\begin{theorem}
\label{thm:finite_cost_separation}
Under Assumption~\ref{ass:1}, for any
\begin{equation}
\label{eq:compatible_cost_interval}
c_1
\in
\left[
\max\{0,\underline{c}_1\},
\infty
\right)
\cap
\mathcal{I}_c,
\end{equation}
the learned CBF satisfies
\begin{equation}
\label{eq:simultaneous_classification}
\begin{aligned}
B(\bm{x}_k)
&\leq-\varepsilon_u<0,
&&
\forall\,\bm{x}_k\in\mathcal{X}_u,\\
B(\bm{x}_c)
&\geq\varepsilon_s\geq0.
\end{aligned}
\end{equation}

Therefore, all unsafe training samples belong to the learned
unsafe set, while the current state belongs to the learned
safe set.
\end{theorem}

\begin{pf}
Define
\(
\bm{R}
:=
\bm{Y}(\bm{I}-\bm{C}).
\)
For unsafe and safe samples, the corresponding rows of
$\bm{R}$ are $[1,-c_1]$ and $[-c_2,1]$, respectively.
From Eq.~\eqref{eq:W_b_we},
\(
\widehat{\bm{Y}}
=
\bm{H}\bm{R}.
\)
Hence, for any sample $\bm{x}_k$,
\begin{equation}
\label{eq:prediction_aggregate}
\begin{aligned}
\widehat{Y}_1(\bm{x}_k)
&=
S_k-c_2T_k,\\
\widehat{Y}_2(\bm{x}_k)
&=
-c_1S_k+T_k,
\end{aligned}
\end{equation}
and therefore
\begin{equation}
\label{eq:B_aggregate}
B(\bm{x}_k)
=
-(1+c_1)S_k
+
(1+c_2)T_k.
\end{equation}

For any $\bm{x}_k\in\mathcal{X}_u$,
$c_1\geq\underline{c}_1$ and $S_k>0$ imply
\[
(1+c_1)S_k
\geq
(1+c_2)T_k+\varepsilon_u,
\]
and thus
$B(\bm{x}_k)\leq-\varepsilon_u$.
Moreover, $c_1\in\mathcal{I}_c$ directly gives
$B(\bm{x}_c)\geq\varepsilon_s$ from
Eqs.~\eqref{eq:current_cost_set} and
\eqref{eq:B_aggregate}.
Finally, Eq.~\eqref{eq:cost_compatibility} guarantees the
existence of a nonnegative $c_1$ satisfying both conditions,
which completes the proof.
\end{pf}

The compatible cost set in
Theorem~\ref{thm:finite_cost_separation} provides a criterion
for selecting the cost $c_1$. When the current
cost parameter violates this compatibility condition, it can be
adapted without retraining the RVFL model from scratch. Since
changing $c_1$ only modifies the cost matrix while leaving the
feature matrix unchanged, the induced variation of the output
weights and CBF can be computed analytically as follows.

\begin{theorem}
\label{thm:c1_update}
For a fixed training set, consider a change of the
misclassification cost from $c_1$ to
$c_{1,\mathrm{new}}=c_1+\Delta c_1\geq0$,
where $\Delta c_1\in\mathbb{R}$ and $c_2$ remains unchanged.
Then the updated output weight matrix satisfies
\begin{equation}
\label{eq:c1_weight_update}
\bm{W}_{b,\mathrm{new}}
=
\bm{W}_b
+
\Delta c_1
\bm{K}\bm{A}_u^\top\bm{1}_u
\begin{bmatrix}
0 & -1
\end{bmatrix},
\end{equation}
and the corresponding updated CBF satisfies
\begin{equation}
\label{eq:c1_barrier_update}
B_{\mathrm{new}}(\bm{x})
=
B(\bm{x})
-
\Delta c_1
\widetilde{\bm{x}}^\top
\bm{K}\bm{A}_u^\top\bm{1}_u,
\end{equation}
where $\bm{A}_u$ contains the extended feature vectors of the
unsafe training samples, $\bm{1}_u$ is an all-ones vector of
the corresponding dimension, and
$\bm{K}=(\lambda\bm{I}+\bm{A}^\top\bm{A})^{-1}$.
\end{theorem}

\begin{pf}
Let
\[
\Delta\bm{C}
=
\begin{bmatrix}
0 & \Delta c_1\\
0 & 0
\end{bmatrix}.
\]

Since
$\bm{C}_{\mathrm{new}}=\bm{C}+\Delta\bm{C}$,
the updated matrix $\bm{Q}_{\mathrm{new}}$ satisfies
\[
\bm{Q}_{\mathrm{new}}
=
\bm{A}^\top\bm{Y}
(\bm{I}-\bm{C}_{\mathrm{new}})
=
\bm{Q}
-
\bm{A}^\top\bm{Y}\Delta\bm{C}.
\]
Only the unsafe labels $[1,0]^\top$ contribute to
$\bm{Y}\Delta\bm{C}$, and hence
\[
\bm{Q}_{\mathrm{new}}
=
\bm{Q}
+
\Delta c_1
\bm{A}_u^\top\bm{1}_u
\begin{bmatrix}
0 & -1
\end{bmatrix}.
\]

Since the feature matrix and regularization parameter remain
unchanged, it holds that
\[
\bm{K}_{\mathrm{new}}
=
(\lambda\bm{I}+\bm{A}^\top\bm{A})^{-1}
=
\bm{K}.
\]

Therefore,
\(
\bm{W}_{b,\mathrm{new}}
=
\bm{K}\bm{Q}_{\mathrm{new}}
\)
gives Eq.~\eqref{eq:c1_weight_update}. Furthermore, since
$[0,-1]\bm{v}_B=-1$, substituting
Eq.~\eqref{eq:c1_weight_update} into the CBF expression yields
Eq.~\eqref{eq:c1_barrier_update}.
\end{pf}

Thus, after selecting a compatible $c_1$, the corresponding
output weights and CBF can be obtained through the analytical
adaptation in Theorem~\ref{thm:c1_update}, without recomputing
$\bm{K}$ or retraining the RVFL model. Under
Theorem~\ref{thm:finite_cost_separation}, the prescribed CBF
margins then hold at the training samples, while
Corollary~\ref{cor:generalization} extends this result to their
neighborhoods.

\begin{corollary}
\label{cor:generalization}
Under the condition of Theorem~\ref{thm:finite_cost_separation},
where
$B(\bm{x}_k)\leq-\varepsilon_u$
for all $\bm{x}_k\in\mathcal X_u$, let $L_B>0$ be a Lipschitz constant of the learned CBF
$B(\bm{x})$. For each unsafe training sample
$\bm{x}_k\in\mathcal{X}_u$, define
$\delta_k:=-B(\bm{x}_k)/L_B
\geq\varepsilon_u/L_B>0$ and
$\mathcal{B}(\bm{x}_k,\delta_k)
:=\{\bm{x}\in\mathbb{R}^n:
\|\bm{x}-\bm{x}_k\|_2<\delta_k\}$.
Then
$\mathcal{B}(\bm{x}_k,\delta_k)
\subseteq\widehat{\mathcal{R}}_u$
for every $\bm{x}_k\in\mathcal{X}_u$.

Moreover, define the covering radius of the unsafe samples over
the real unsafe region as
\begin{equation}
\label{eq:unsafe_covering_radius}
\rho_u
:=
\sup_{\bm{x}\in\mathcal{R}_u}
\min_{\bm{x}_k\in\mathcal{X}_u}
\|\bm{x}-\bm{x}_k\|_2,
\end{equation}
and let
$\delta_{\min}:=
\min_{\bm{x}_k\in\mathcal{X}_u}\delta_k$.
If $\rho_u<\delta_{\min}$, then
\begin{equation}
\label{eq:conservative_unsafe_coverage}
\mathcal{R}_u
\subseteq
\bigcup_{\bm{x}_k\in\mathcal{X}_u}
\mathcal{B}(\bm{x}_k,\delta_k)
\subseteq
\widehat{\mathcal{R}}_u.
\end{equation}
\end{corollary}

\begin{pf}
For any
$\bm{x}\in\mathcal{B}(\bm{x}_k,\delta_k)$, the Lipschitz
continuity of $B$ gives
\begin{align}
B(\bm{x})
&\leq
B(\bm{x}_k)
+
L_B\|\bm{x}-\bm{x}_k\|_2
\nonumber\\
&<
B(\bm{x}_k)+L_B\delta_k
=
0.
\end{align}

Hence,
$\mathcal{B}(\bm{x}_k,\delta_k)
\subseteq\widehat{\mathcal{R}}_u$.

If $\rho_u<\delta_{\min}$, then for any
$\bm{x}\in\mathcal{R}_u$, there exists
$\bm{x}_k\in\mathcal{X}_u$ such that
$\|\bm{x}-\bm{x}_k\|_2
\leq\rho_u<\delta_{\min}\leq\delta_k$.
Thus,
$\bm{x}\in\mathcal{B}(\bm{x}_k,\delta_k)
\subseteq\widehat{\mathcal{R}}_u$.
Since $\bm{x}$ is arbitrary,
Eq.~\eqref{eq:conservative_unsafe_coverage} follows.
\end{pf}

Corollary~\ref{cor:generalization} provides a sufficient
condition for conservatively covering the real unsafe region
when $\rho_u$, or a valid upper bound of it, is available.
Since updating the training set changes the learned CBF and
therefore the corresponding Lipschitz constant $L_B$ and
coverage radii $\delta_k$, the covering condition
$\rho_u<\delta_{\min}$ is re-evaluated after each update.

Combining the conservative inclusion
$\mathcal{R}_{u,\tau}\subseteq\widehat{\mathcal{R}}_{u,\tau}$
with the HOCBF forward-invariance property yields the following
interval-wise safety guarantee.

\begin{corollary}
\label{cor:subset_safety}
Consider an interval $[t_\tau,t_{\tau+1})$ over which the real
unsafe region $\mathcal{R}_{u,\tau}$ and the learned CBF
$B_\tau(\bm{x})$ remain unchanged. Suppose that
$\mathcal{R}_{u,\tau}\subseteq
\widehat{\mathcal{R}}_{u,\tau}$
and
$\bm{x}(t_\tau)\in
\bigcap_{i=1}^{r}\widehat{\mathcal{C}}_{i,\tau}$,
where $\widehat{\mathcal{C}}_{i,\tau}$ denotes the HOCBF set
associated with $B_\tau$ according to
Eqs.~(\ref{eq:psi}) and (\ref{eq:psi_C}).
If the corresponding HOCBF-QP remains feasible and the resulting control input $\bm{u}_{\mathrm{safe}}$
is Lipschitz continuous on
$[t_\tau,t_{\tau+1})$, then
$\bm{x}(t)\notin\mathcal{R}_{u,\tau}$ for all
$t\in[t_\tau,t_{\tau+1})$.
\end{corollary}

\begin{pf}
Under the stated conditions, the HOCBF forward-invariance
result gives
$\bm{x}(t)\in
\bigcap_{i=1}^{r}\widehat{\mathcal{C}}_{i,\tau}
\subseteq\widehat{\mathcal{C}}_{1,\tau}$
for all $t\in[t_\tau,t_{\tau+1})$.
Since
$\mathcal{R}_{u,\tau}
\subseteq\widehat{\mathcal{R}}_{u,\tau}$,
taking complements gives
$\widehat{\mathcal{C}}_{1,\tau}
=
\mathbb{R}^n\setminus\widehat{\mathcal{R}}_{u,\tau}
\subseteq
\mathbb{R}^n\setminus\mathcal{R}_{u,\tau}$.
Therefore,
$\bm{x}(t)\notin\mathcal{R}_{u,\tau}$ for all
$t\in[t_\tau,t_{\tau+1})$, which completes the proof.
\end{pf}

\subsection{SafeLink Adaptation to Evolving Unsafe Regions}
\label{subsec:safelink_update}

We consider a sequence of update times
$\{t_\tau\}_{\tau\in\mathbb{Z}_{\geq0}}$, where the real
unsafe region and the learned CBF remain unchanged within each
interval $[t_\tau,t_{\tau+1})$. At each update time
$t_{\tau+1}$, the training set is adapted according to the
latest safety information: obsolete samples are removed, newly
observed samples are incorporated, and samples with changed
safety labels are updated accordingly. The current
state is additionally included as a safe sample with label
$[0,1]^\top$.

To efficiently accommodate these changes, we derive
decremental and incremental update rules for SafeLink.
The proposed analytical updates preserve the exact solution of
the RVFL optimization problem: in exact arithmetic, they are
algebraically equivalent to retraining the model on the
corresponding updated dataset.

For the model associated with the $\tau$-th interval, define
\begin{equation}
\label{eq:update_KQ}
\begin{aligned}
\bm{K}_\tau
&:=
\left(
\lambda\bm{I}
+\bm{A}_\tau^\top\bm{A}_\tau
\right)^{-1},\\
\bm{Q}_\tau
&:=
\bm{A}_\tau^\top\bm{Y}_\tau(\bm{I}-\bm{C}),\\
B_\tau(\bm{x})
&:=
\widetilde{\bm{x}}^\top
\bm{K}_\tau\bm{Q}_\tau\bm{v}_B.
\end{aligned}
\end{equation}

At $t_{\tau+1}$, let $\Delta N^-$ and $\Delta N^+$ denote
the numbers of removed and added samples, respectively. The
corresponding feature and label matrices are denoted by
\begin{equation}
\label{eq:changed_samples}
\begin{aligned}
\Delta\bm{A}_{\tau+1}^-
&\in
\mathbb{R}^{\Delta N^-\times(n+M)},&
\Delta\bm{Y}_{\tau+1}^-
&\in
\mathbb{R}^{\Delta N^-\times2},\\
\Delta\bm{A}_{\tau+1}^+
&\in
\mathbb{R}^{\Delta N^+\times(n+M)},&
\Delta\bm{Y}_{\tau+1}^+
&\in
\mathbb{R}^{\Delta N^+\times2}.
\end{aligned}
\end{equation}

Here,
$(\Delta\bm{A}_{\tau+1}^-,
\Delta\bm{Y}_{\tau+1}^-)$ represents the
samples to be removed, whereas
$(\Delta\bm{A}_{\tau+1}^+,
\Delta\bm{Y}_{\tau+1}^+)$
represents the samples to be added. For a relabeled sample,
its previous version is included in the former and its updated
version in the latter. When the current state is confirmed to
lie outside $\mathcal{R}_{u,\tau+1}$, its extended feature
vector and safe label $[0,1]^\top$ can be included in the
added samples. The decremental and incremental operations can be
applied separately or sequentially as needed.

The decremental update rule for removing obsolete samples is
derived as follows.

\begin{theorem}
\label{thm:decremental_update}
Consider the model in Eq.~\eqref{eq:update_KQ} and suppose that
the samples represented by
$(\Delta\bm{A}_{\tau+1}^-,
\Delta\bm{Y}_{\tau+1}^-)$ are removed.
Define
\begin{equation}
\label{eq:removal_deltas}
\begin{aligned}
\Delta\bm{K}_{\tau+1}^-
&:=
\bm{K}_\tau
(\Delta\bm{A}_{\tau+1}^-)^\top
\Big(
\bm{I}
-
\Delta\bm{A}_{\tau+1}^-\bm{K}_\tau
(\Delta\bm{A}_{\tau+1}^-)^\top
\Big)^{-1}\\
&\hspace{2.5em}\times
\Delta\bm{A}_{\tau+1}^-\bm{K}_\tau,\\
\Delta\bm{Q}_{\tau+1}^-
&:=
(\Delta\bm{A}_{\tau+1}^-)^\top
\Delta\bm{Y}_{\tau+1}^-(\bm{I}-\bm{C}).
\end{aligned}
\end{equation}
Then the decrementally updated matrices are
\begin{equation}
\label{eq:removal_KQ}
\bm{K}_{\tau+1}^{\mathrm{dec}}
=
\bm{K}_\tau+\Delta\bm{K}_{\tau+1}^-,
\qquad
\bm{Q}_{\tau+1}^{\mathrm{dec}}
=
\bm{Q}_\tau-\Delta\bm{Q}_{\tau+1}^-,
\end{equation}
and the corresponding CBF satisfies
\begin{equation}
\label{eq:removal_barrier}
\begin{aligned}
B_{\tau+1}^{\mathrm{dec}}(\bm{x})
=
B_\tau(\bm{x})
+
\widetilde{\bm{x}}^\top
\Big(
&-\bm{K}_\tau\Delta\bm{Q}_{\tau+1}^-
+\Delta\bm{K}_{\tau+1}^-\bm{Q}_\tau\\
&-\Delta\bm{K}_{\tau+1}^-
\Delta\bm{Q}_{\tau+1}^-
\Big)\bm{v}_B.
\end{aligned}
\end{equation}
\end{theorem}

\begin{pf}
Let $\bm{A}_{\tau+1}^{\mathrm{dec}}$ denote the feature matrix
after removing the specified samples. Then
\[
(\bm{A}_{\tau+1}^{\mathrm{dec}})^\top
\bm{A}_{\tau+1}^{\mathrm{dec}}
=
\bm{A}_\tau^\top\bm{A}_\tau
-
(\Delta\bm{A}_{\tau+1}^-)^\top
\Delta\bm{A}_{\tau+1}^-.
\]
Applying the Woodbury matrix identity
\cite{hager1989updating,li2025dynamic} gives
$\bm{K}_{\tau+1}^{\mathrm{dec}}
=\bm{K}_\tau+\Delta\bm{K}_{\tau+1}^-$. Similarly,
$\bm{Q}_{\tau+1}^{\mathrm{dec}}
=\bm{Q}_\tau-\Delta\bm{Q}_{\tau+1}^-$. Expanding their product
and substituting it into the CBF expression yields
Eq.~\eqref{eq:removal_barrier}.
\end{pf}

The incremental update rule for incorporating newly observed
samples is derived similarly.

\begin{theorem}
\label{thm:incremental_update}
Consider the model in Eq.~\eqref{eq:update_KQ} and suppose that
the samples represented by
$(\Delta\bm{A}_{\tau+1}^+,\Delta\bm{Y}_{\tau+1}^+)$ are added.
Define
\begin{equation}
\label{eq:addition_deltas}
\begin{aligned}
\Delta\bm{K}_{\tau+1}^+
&:=
\bm{K}_\tau
(\Delta\bm{A}_{\tau+1}^+)^\top
\Big(
\bm{I}
+
\Delta\bm{A}_{\tau+1}^+\bm{K}_\tau
(\Delta\bm{A}_{\tau+1}^+)^\top
\Big)^{-1}\\
&\hspace{2.5em}\times
\Delta\bm{A}_{\tau+1}^+\bm{K}_\tau,\\
\Delta\bm{Q}_{\tau+1}^+
&:=
(\Delta\bm{A}_{\tau+1}^+)^\top
\Delta\bm{Y}_{\tau+1}^+(\bm{I}-\bm{C}).
\end{aligned}
\end{equation}
Then the incrementally updated matrices are
\begin{equation}
\label{eq:addition_KQ}
\bm{K}_{\tau+1}^{\mathrm{inc}}
=
\bm{K}_\tau-\Delta\bm{K}_{\tau+1}^+,
\qquad
\bm{Q}_{\tau+1}^{\mathrm{inc}}
=
\bm{Q}_\tau+\Delta\bm{Q}_{\tau+1}^+,
\end{equation}
and the corresponding CBF satisfies
\begin{equation}
\label{eq:incremental_barrier}
\begin{aligned}
B_{\tau+1}^{\mathrm{inc}}(\bm{x})
=
B_\tau(\bm{x})
+
\widetilde{\bm{x}}^\top
\Big(
&\bm{K}_\tau\Delta\bm{Q}_{\tau+1}^+
-\Delta\bm{K}_{\tau+1}^+\bm{Q}_\tau\\
&-\Delta\bm{K}_{\tau+1}^+
\Delta\bm{Q}_{\tau+1}^+
\Big)\bm{v}_B.
\end{aligned}
\end{equation}
\end{theorem}

\begin{pf}
Let $\bm{A}_{\tau+1}^{\mathrm{inc}}$ denote the feature matrix
after adding the specified samples. Then
\[
\bm{A}_{\tau+1}^{\mathrm{inc}}
=
\begin{bmatrix}
\bm{A}_\tau\\
\Delta\bm{A}_{\tau+1}^+
\end{bmatrix},
\]
and therefore
\[
(\bm{A}_{\tau+1}^{\mathrm{inc}})^\top
\bm{A}_{\tau+1}^{\mathrm{inc}}
=
\bm{A}_\tau^\top\bm{A}_\tau
+
(\Delta\bm{A}_{\tau+1}^+)^\top
\Delta\bm{A}_{\tau+1}^+.
\]

Applying the Woodbury matrix identity gives
$\bm{K}_{\tau+1}^{\mathrm{inc}}
=\bm{K}_\tau-\Delta\bm{K}_{\tau+1}^+$,
while direct expansion gives
$\bm{Q}_{\tau+1}^{\mathrm{inc}}
=\bm{Q}_\tau+\Delta\bm{Q}_{\tau+1}^+$.
Expanding their product and substituting it into the CBF
expression yields Eq.~\eqref{eq:incremental_barrier}.
\end{pf}

If both decremental and incremental updates are required at the
same update time, they can be applied in either order, with the
updated matrices from one operation used as the input to the
other. Since both updates are exact algebraic operations, their
composition is equivalent, in exact arithmetic, to batch
retraining on the resulting dataset.

The decremental and incremental updates require matrix inverses
of dimensions $\Delta N^-$ and $\Delta N^+$, respectively.
Thus, when $\Delta N^-$ and $\Delta N^+$ are small compared
with the feature dimension $n+M$, these updates avoid
recomputing the full
$(n+M)\times(n+M)$ inverse.

After each data update, Assumption~\ref{ass:1} and the
compatible cost set in
Theorem~\ref{thm:finite_cost_separation} are re-evaluated. If
the compatibility condition remains satisfied but the
previously selected $c_1$ no longer belongs to the updated
compatible cost set, $c_1$ and the corresponding CBF are
adjusted using
Theorem~\ref{thm:c1_update} without retraining the RVFL model.
If the compatibility condition itself is violated, adjusting
$c_1$ alone cannot recover the guarantees of
Theorem~\ref{thm:finite_cost_separation}.

\begin{rmk}
\label{rmk:dynamic_safety}
After each update, suppose that 
Assumption~\ref{ass:1} holds. If the previously selected 
$c_1$ no longer belongs to the updated compatible cost set, it 
is adjusted according to Theorem~\ref{thm:c1_update} to recover 
the condition in Theorem~\ref{thm:finite_cost_separation}. For 
systems with $r=1$, the SafeLink adaptation adds the current state as a safe 
sample, and Theorem~\ref{thm:finite_cost_separation} ensures that
\(
\bm{x}(t_{\tau+1})\in
\widehat{\mathcal{C}}_{1,\tau+1}.
\)
Therefore, Corollary~\ref{cor:subset_safety} guarantees safety 
over the subsequent update interval under the stated HOCBF-QP 
feasibility and local Lipschitz conditions. For systems with $r>1$, the 
additional condition
\(
\bm{x}(t_{\tau+1})\in
\bigcap_{i=2}^{r}\widehat{\mathcal{C}}_{i,\tau+1}
\)
is required for the same guarantee.
\end{rmk}
\color{blue}

\color{black}

\section{Experiments}
\label{sec:experiments}
This section evaluates SafeLink on a two-link manipulator under evolving irregular unsafe regions with sudden expansions. The experiments examine the effect of the misclassification cost, the adaptation performance under evolving unsafe-region changes, and the computational efficiency relative to other learning-based CBF methods.

\subsection{Settings}
A two-link manipulator with second-order joint dynamics and nonlinear endpoint kinematics is considered, with its base fixed at the origin. The link lengths are $L_1 = L_2 = 4$ m. The angle between the first link and the $x$-axis is denoted by $\theta_1$ with angular velocity $\omega_1$, while the angle between the second link and the first link is denoted by $\theta_2$ with angular velocity $\omega_2$. The system states and dynamics are defined as:
\begin{equation}
\left\{
\begin{aligned}
\bm{z} &=
\begin{bmatrix}
\theta_1, 
\theta_2,
\omega_1,
\omega_2
\end{bmatrix}^\top, \\[2pt]
\dot{\bm{z}} &=
\begin{bmatrix}
0 & 0 & 1 & 0 \\
0 & 0 & 0 & 1 \\
0 & 0 & 0 & 0 \\
0 & 0 & 0 & 0
\end{bmatrix}
\begin{bmatrix}
\theta_1 \\[2pt]
\theta_2 \\[2pt]
\omega_1 \\[2pt]
\omega_2
\end{bmatrix}
+
\begin{bmatrix}
0 & 0 \\
0 & 0 \\
1 & 0 \\
0 & 1
\end{bmatrix}
\begin{bmatrix}
u_1 \\[2pt]
u_2
\end{bmatrix}, \\[2pt]
x &= L_1\cos\theta_1 + L_2\cos(\theta_1+\theta_2), \\
y &= L_1\sin\theta_1 + L_2\sin(\theta_1+\theta_2),
\end{aligned}
\right.
\end{equation}
where $(x,y)$ denotes the endpoint position, and the control inputs are constrained as $u_1, u_2 \in [-2,2]$ rad/s\textsuperscript{2}.

The RVFL is configured with $M = 100$, $\lambda = 0.001$, and cost-sensitivity parameter $c_2 = 1$. The activation function is chosen as $\phi(x)=\operatorname{sigmoid}(5x)$, whose derivatives required by the second-order HOCBF are Lipschitz continuous. Both $\alpha_2$ in Eq.~(\ref{eq:u_HOCBF}) and $\alpha_1$ in Eq.~(\ref{eq:psi}) are linear functions with unit coefficients. The control interval is fixed at $\Delta t = 0.05$ s.

\begin{figure}[htbp]
    \centering
    \begin{center}
\includegraphics[width=0.5\textwidth]{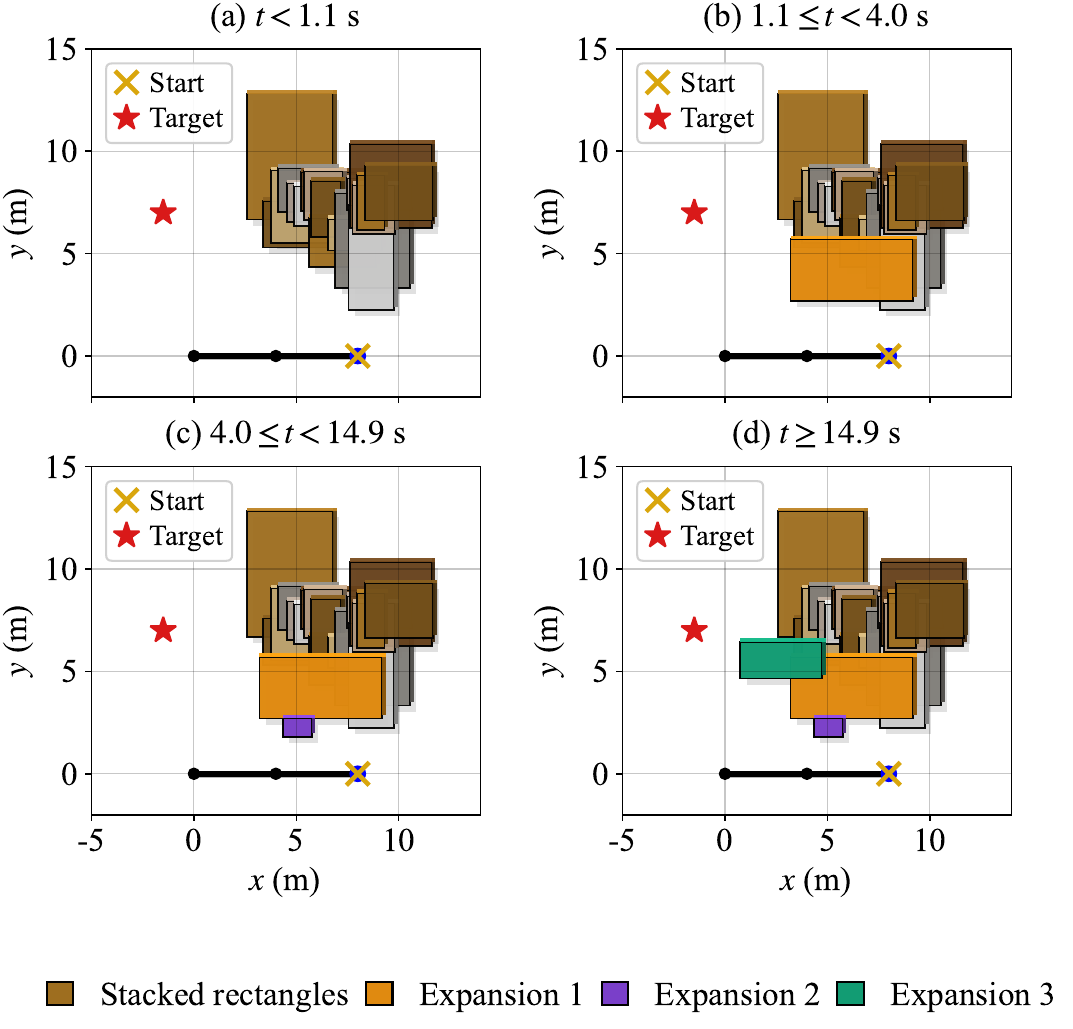}  
    \end{center}
\caption{Evolution of the unsafe region, with sudden expansions at $t=1.1$ s, $t=4.0$ s, and $t=14.9$ s.}
   
\label{fig:region}
\end{figure}

The manipulator is simulated in an environment containing stacked parts and cargo, forming a highly irregular unsafe region that is difficult to represent analytically. The unsafe region is defined as the union of multiple stacked rectangles (Fig.~\ref{fig:region}). In the offline stage, 5,000 endpoint-space samples are uniformly generated within $x\in[-5,15]$ and $y\in[-5,15]$, with additional boundary samples yielding a total of 5,792 labeled samples. Samples outside the reachable workspace are retained to characterize the surrounding unsafe-region geometry and contribute to the learned CBF boundary. Each sample is labeled as $[1,0]^\top$ if unsafe and $[0,1]^\top$ otherwise. The RVFL is trained directly in the endpoint workspace using $[x,y]$ as input, and the learned CBF derivatives are transformed to the manipulator state coordinates through the forward kinematics and the chain rule.

The system is initialized at $[x, y, \theta_1, \theta_2, \omega_1, \omega_2] = [8, 0, 0, 0, 0, 0]$, and the target endpoint position is set to $[x, y] = [-1.5, 7]$ within the reachable workspace. During the simulation, the unsafe region expands three times, at $t = 1.1$ s, $t = 4.0$ s, and $t = 14.9$ s. After each expansion, several dozen new unsafe samples are collected within the expanded region. These samples, together with the current state point of the system, are used for incremental updates, whereas the safe samples that originally lay within the expanded region and the previous state point are removed through decremental updates.

\begin{table*}[htbp]
    \centering
    \caption{Performance of the learned CBF for different \(c_1\) values.}
    \label{tab:c_1}
    \setlength{\tabcolsep}{6mm}{
    \begin{tabular}{@{}l c c c c c c c c@{}}
\specialrule{0.1em}{1pt}{1pt}
\specialrule{0.1em}{1pt}{3pt}
        \multirow{1}{*}{\(c_1\)} & \multirow{1}{*}{1.0} & \multirow{1}{*}{2.0} & \multirow{1}{*}{5.0} & \multirow{1}{*}{5.324} & \multirow{1}{*}{6.92} & \multirow{1}{*}{8.0} & \multirow{1}{*}{10.0}\\
        \specialrule{0.1em}{3pt}{3pt}
          {\(N_{u\to s}\)} & 77 & 19 & 1 & 1 & \textbf{0}& \textbf{0}& \textbf{0} \\
        \specialrule{0.0em}{1pt}{1pt}
         {\(N_{s\to u}\)} & 82 & 157 & 314 & 326 & 387 & 418 & 482\\
\specialrule{0.0em}{1pt}{1pt}
         {\textit{Accuracy}} & 97.25\% & 96.96\% & 94.56\%& 94.35\% & 93.32\% & 92.78\% & 91.68\%\\
\specialrule{0.0em}{1pt}{1pt}
         {\textit{Recall}} & 95.08\% & 98.79\% & 99.94\%& 99.94\% & \textbf{100\%}& \textbf{100\%}& \textbf{100\%}\\
  \specialrule{0.1em}{3pt}{1pt}
\specialrule{0.1em}{1pt}{1pt} 
    \end{tabular}
    }
    \begin{tablenotes}\footnotesize
    \item[1] *Notes: \( N_{s\to u} \) represents the number of safe samples identified as unsafe, and \( N_{u\to  s} \) represents the number of unsafe samples identified as safe. \textit{Accuracy} represents the ratio of correctly identified samples. \textit{Recall} represents the ratio of correctly identified unsafe samples.
\end{tablenotes}
\end{table*}

\begin{figure}[htbp]
    \centering
    \begin{center}
\includegraphics[width=0.4\textwidth]{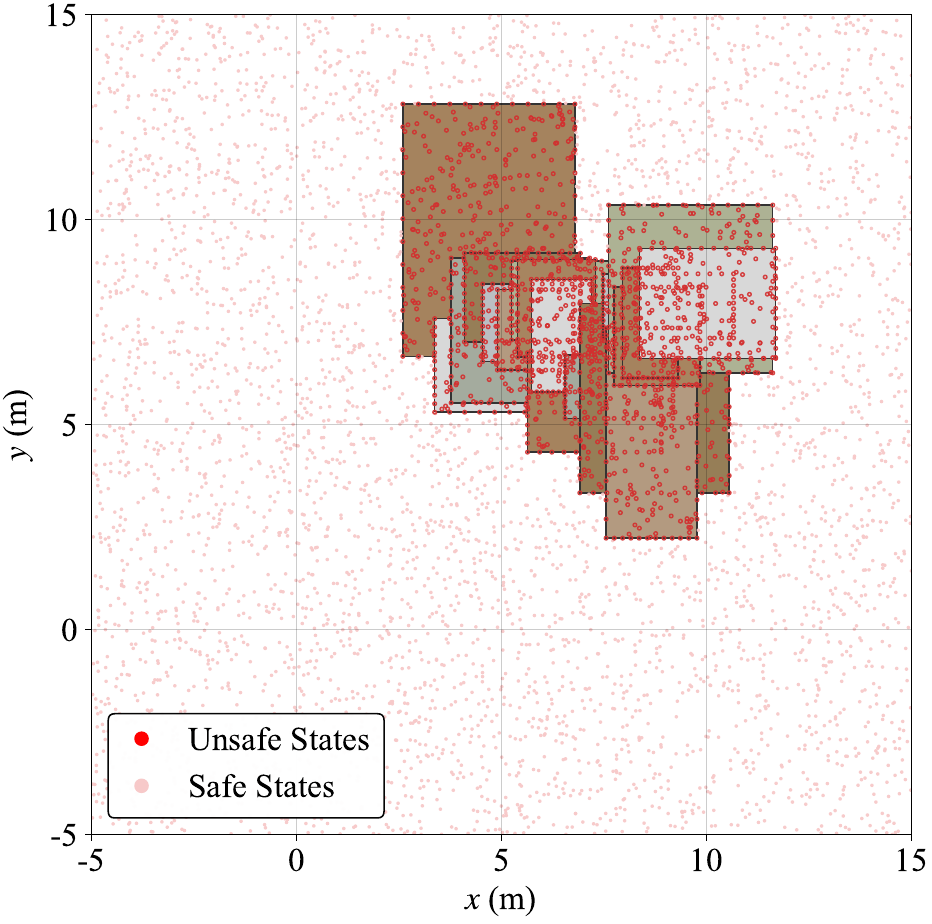}  
    \end{center}
\caption{Sample distribution in the endpoint workspace. A total of 5792 labeled samples are used for offline training.}

\label{fig:samples}
\end{figure}

\subsection{Sensitivity Analysis of Misclassification Cost $c_1$}

This section investigates the impact of the misclassification cost $c_1$ on the classification tradeoff between safe and unsafe samples. Table~\ref{tab:c_1} summarizes the relationship between $c_1$ and the classification performance, which is evaluated on the offline training samples at $t = 0$.

According to Theorem~\ref{thm:finite_cost_separation}, the
compatible cost set at the offline stage is
$c_1\in[6.92,\infty)$. As observed in Table~\ref{tab:c_1},
$N_{u\to s}$ decreases monotonically as $c_1$ increases and
reaches zero for $c_1\geq6.92$, at which point the recall of the
unsafe class attains $100\%$. Increasing $c_1$ reduces
unsafe-to-safe misclassifications at the expense of more
safe-to-unsafe misclassifications, leading to a more conservative
learned unsafe set. Accordingly, $c_1=6.92$ is selected for control, as it 
provides the least conservative CBF while ensuring
$N_{u\to s}=0$ and a $100\%$ recall of the unsafe class.

\begin{figure*}[htbp]
    \centering

    % ==================== Row 1 ====================
    \begin{minipage}[c]{0.33\linewidth}
        \centering
        \includegraphics[width=\linewidth]{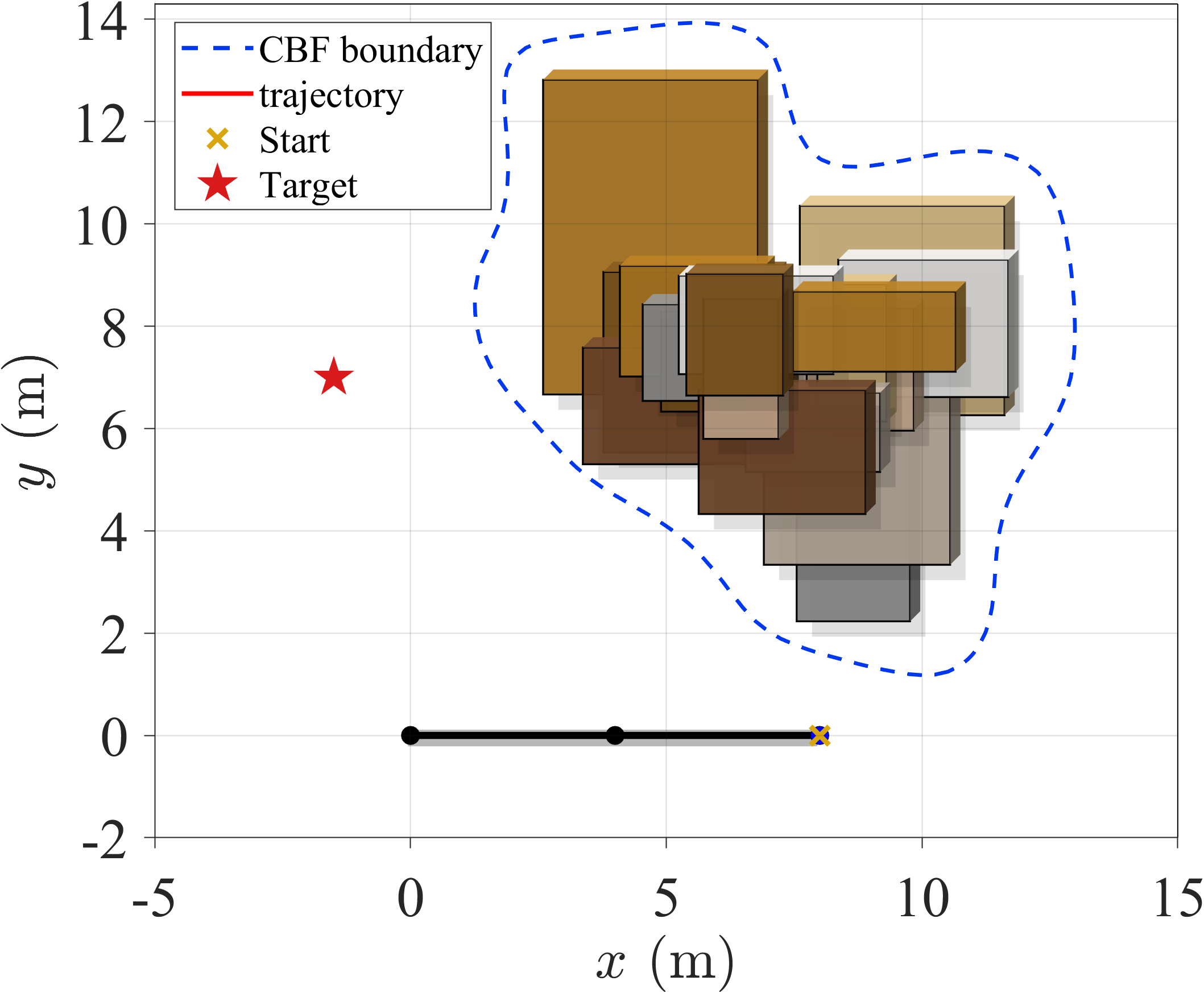}
        \par
        (a) $t=0$ s
    \end{minipage}%
    \hfill
    \begin{minipage}[c]{0.33\linewidth}
        \centering
        \includegraphics[width=\linewidth]{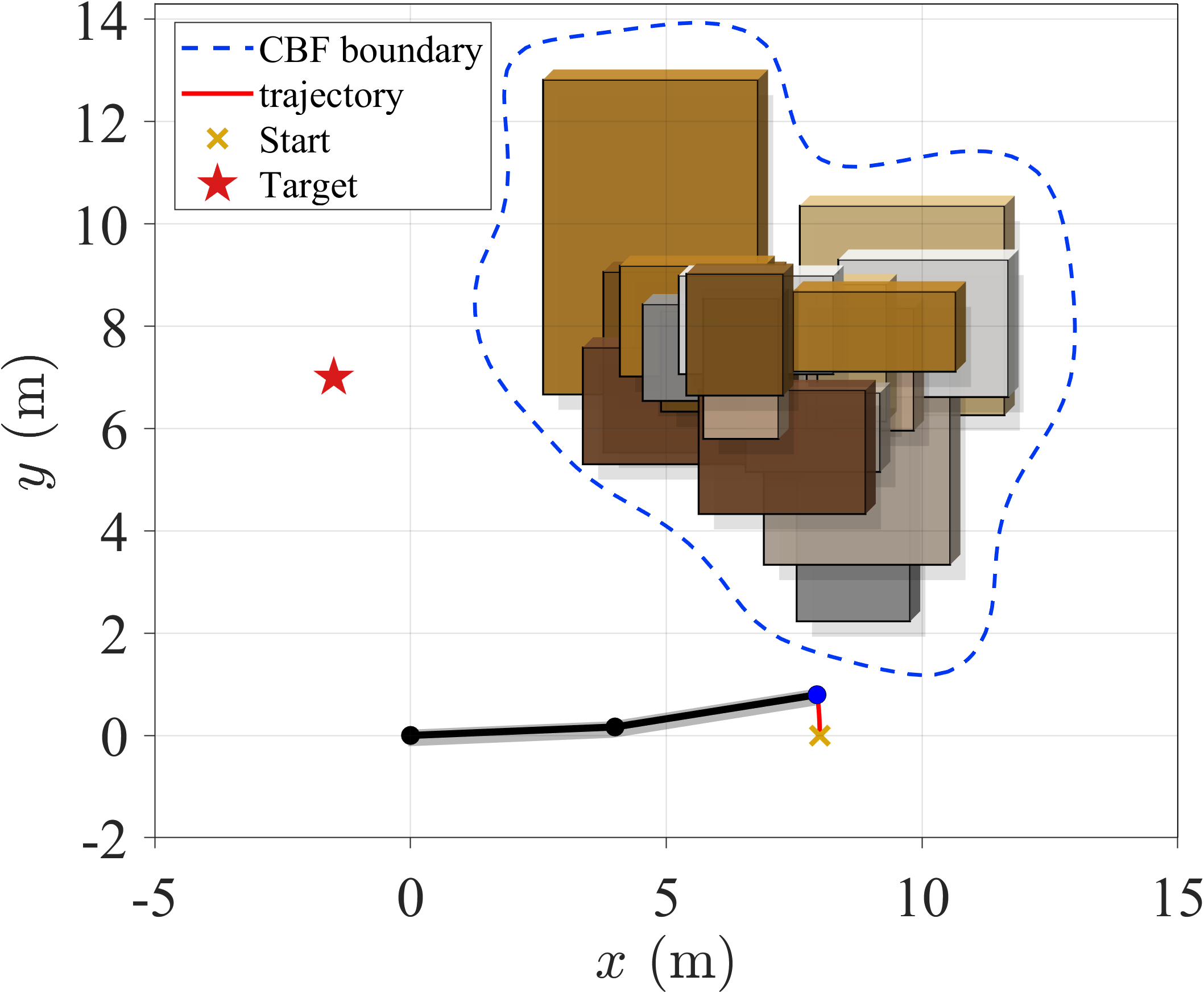}
        \par
        (b) $t=1.05$ s
    \end{minipage}%
    \hfill
    \begin{minipage}[c]{0.33\linewidth}
        \centering
        \includegraphics[width=\linewidth]{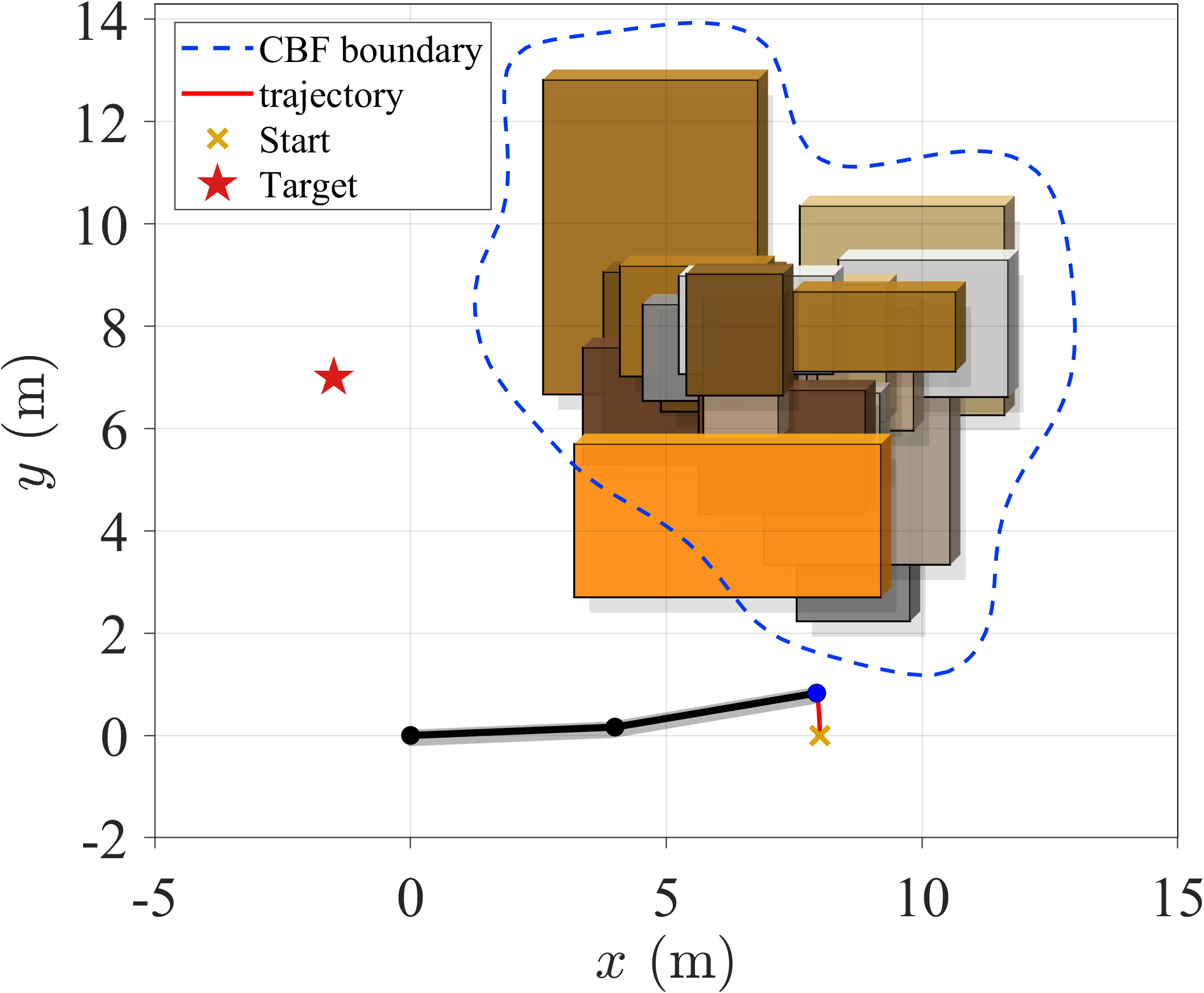}
        \par
        (c) $t=1.10$ s (before update)
    \end{minipage}

    \vspace{4mm}

    % ==================== Row 2 ====================
    \begin{minipage}[c]{0.33\linewidth}
        \centering
        \includegraphics[width=\linewidth]{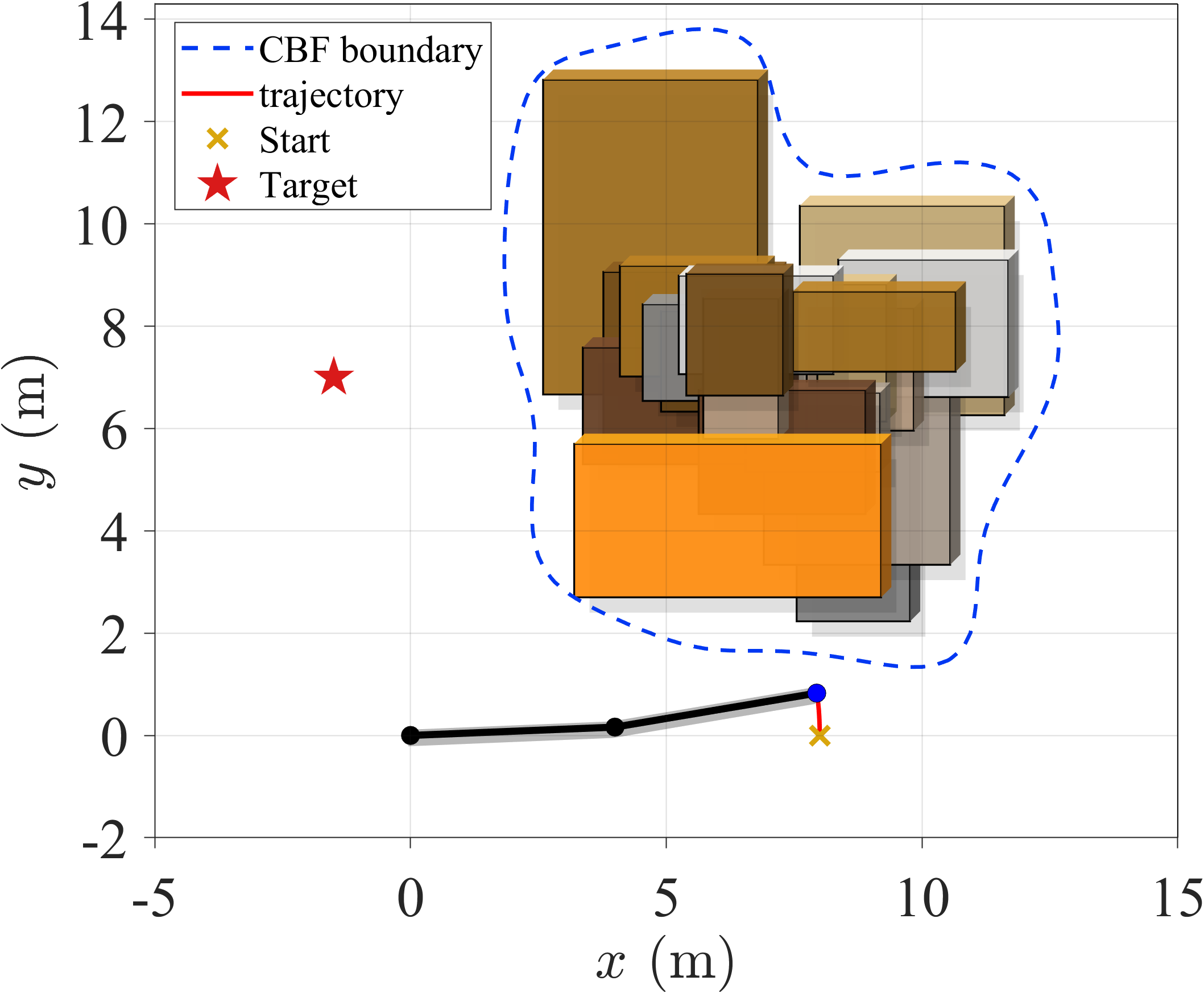}
        \par
        (d) $t=1.10$ s (after update)
    \end{minipage}%
    \hfill
    \begin{minipage}[c]{0.33\linewidth}
        \centering
        \includegraphics[width=\linewidth]{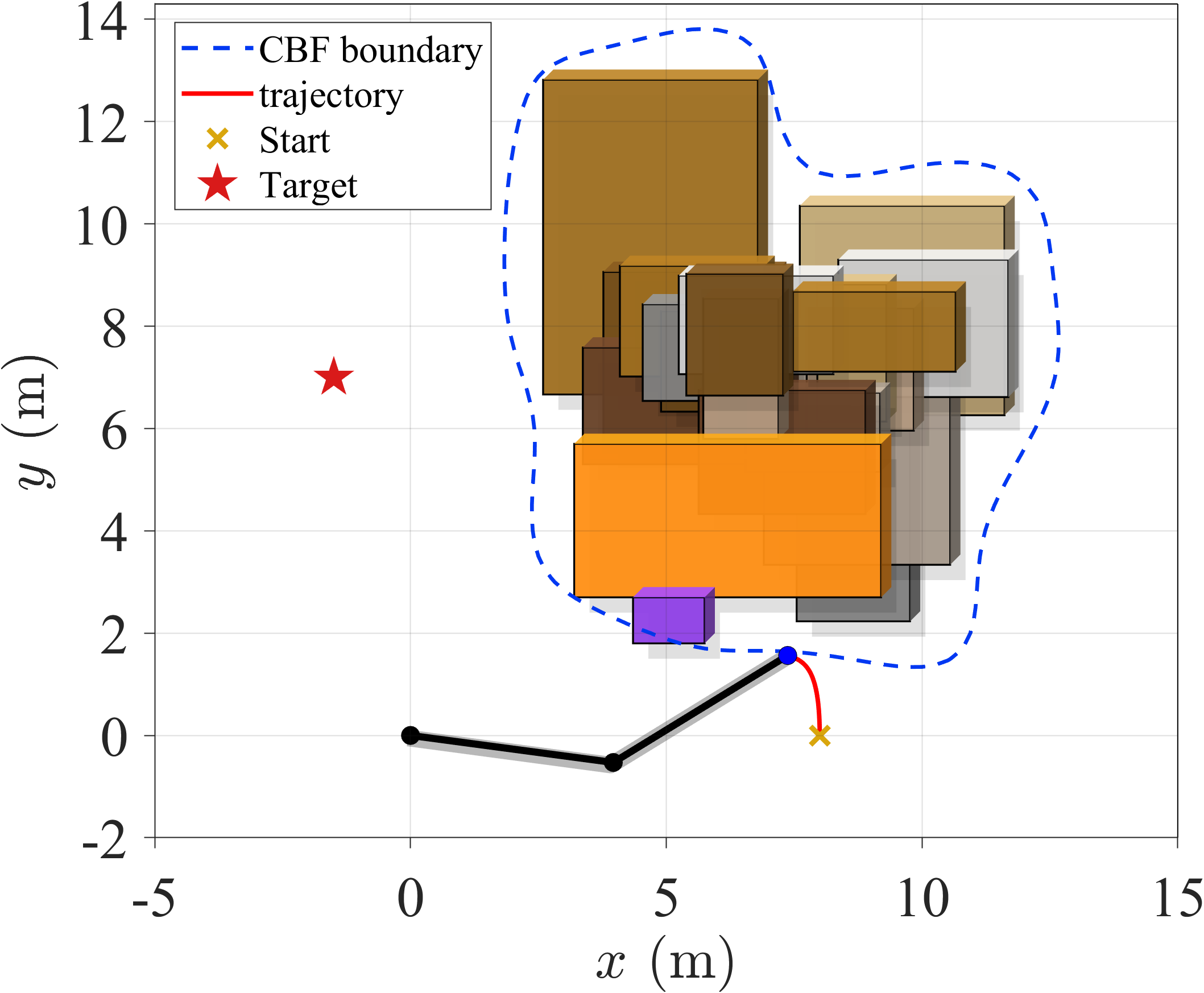}
        \par
        (e) $t=4.00$ s (before update)
    \end{minipage}%
    \hfill
    \begin{minipage}[c]{0.33\linewidth}
        \centering
        \includegraphics[width=\linewidth]{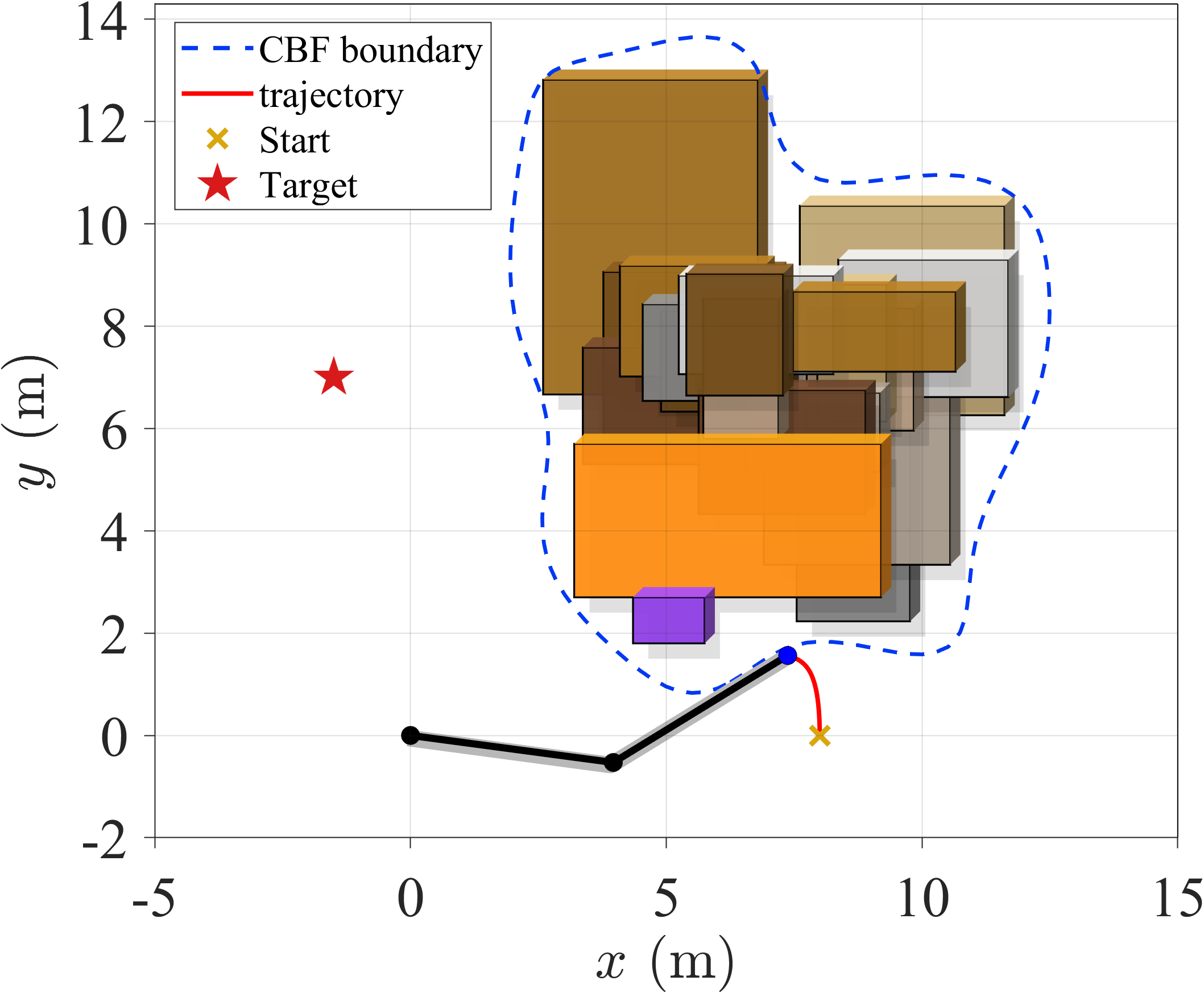}
        \par
        (f) $t=4.00$ s (after update)
    \end{minipage}

    \vspace{4mm}

    % ==================== Row 3 ====================
    \begin{minipage}[c]{0.33\linewidth}
        \centering
        \includegraphics[width=\linewidth]{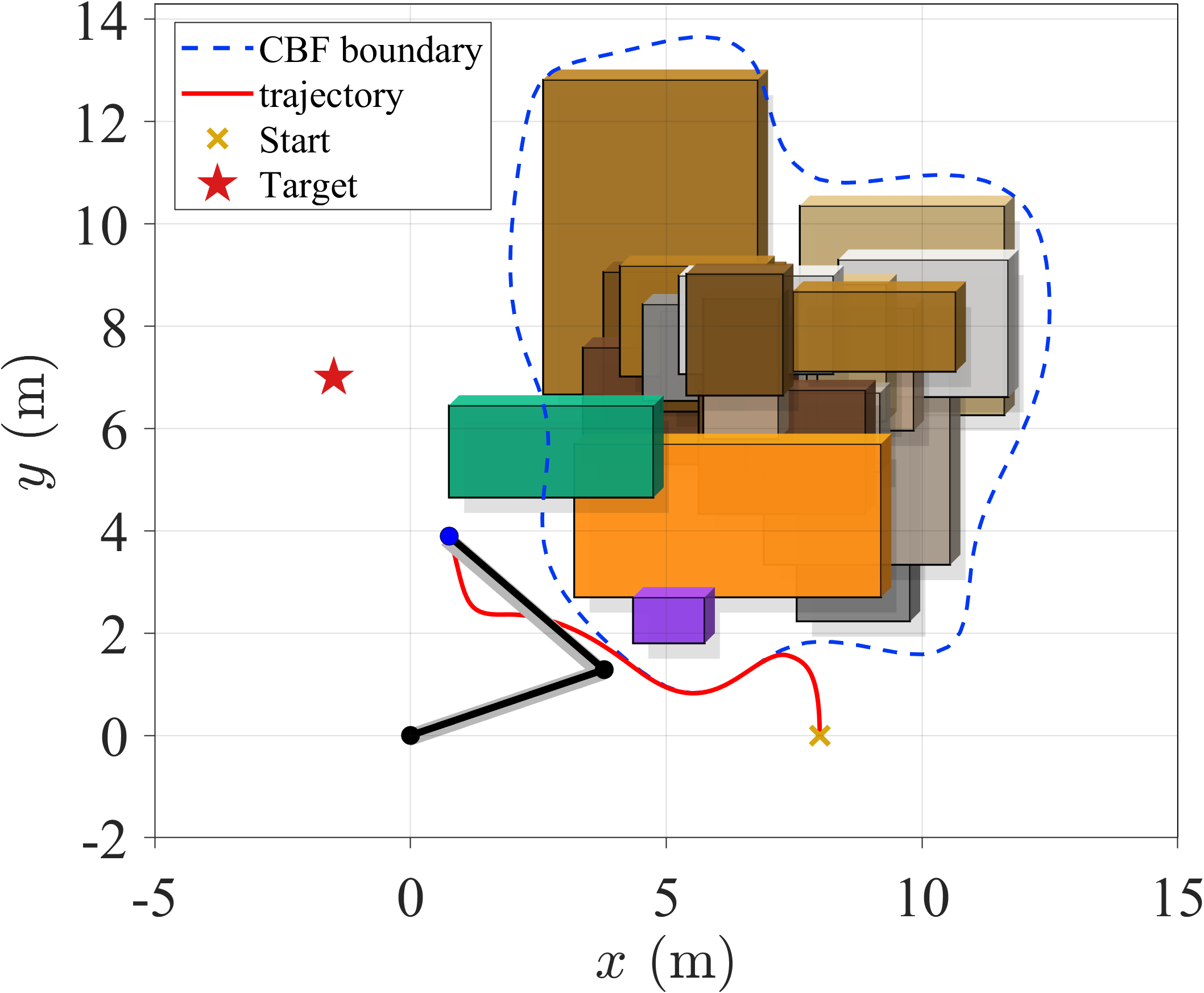}
        \par
        (g) $t=14.90$ s (before update)
    \end{minipage}%
    \hfill
    \begin{minipage}[c]{0.33\linewidth}
        \centering
        \includegraphics[width=\linewidth]{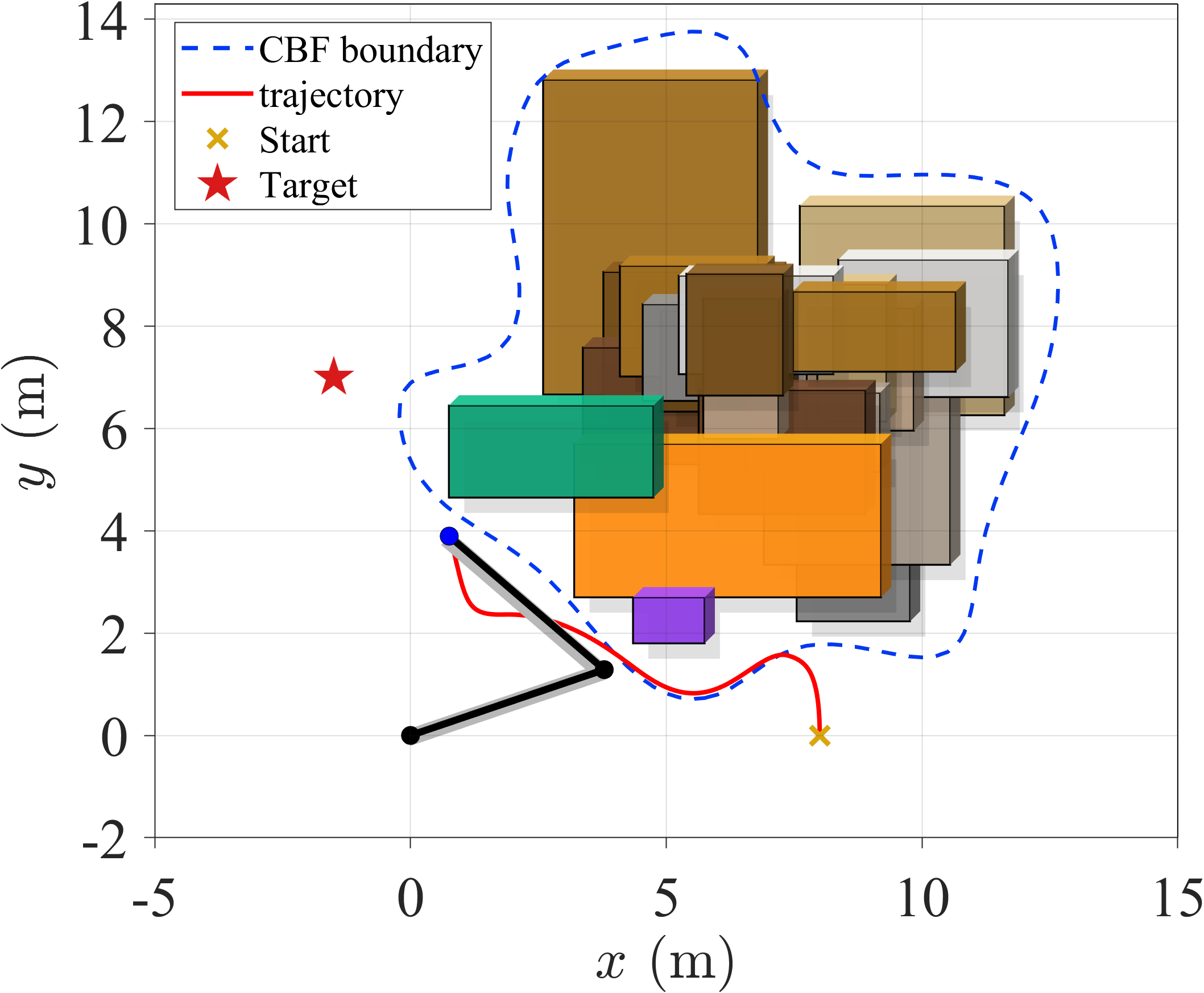}
        \par
        (h) $t=14.90$ s (after update)
    \end{minipage}%
    \hfill
    \begin{minipage}[c]{0.33\linewidth}
        \centering
        \includegraphics[width=\linewidth]{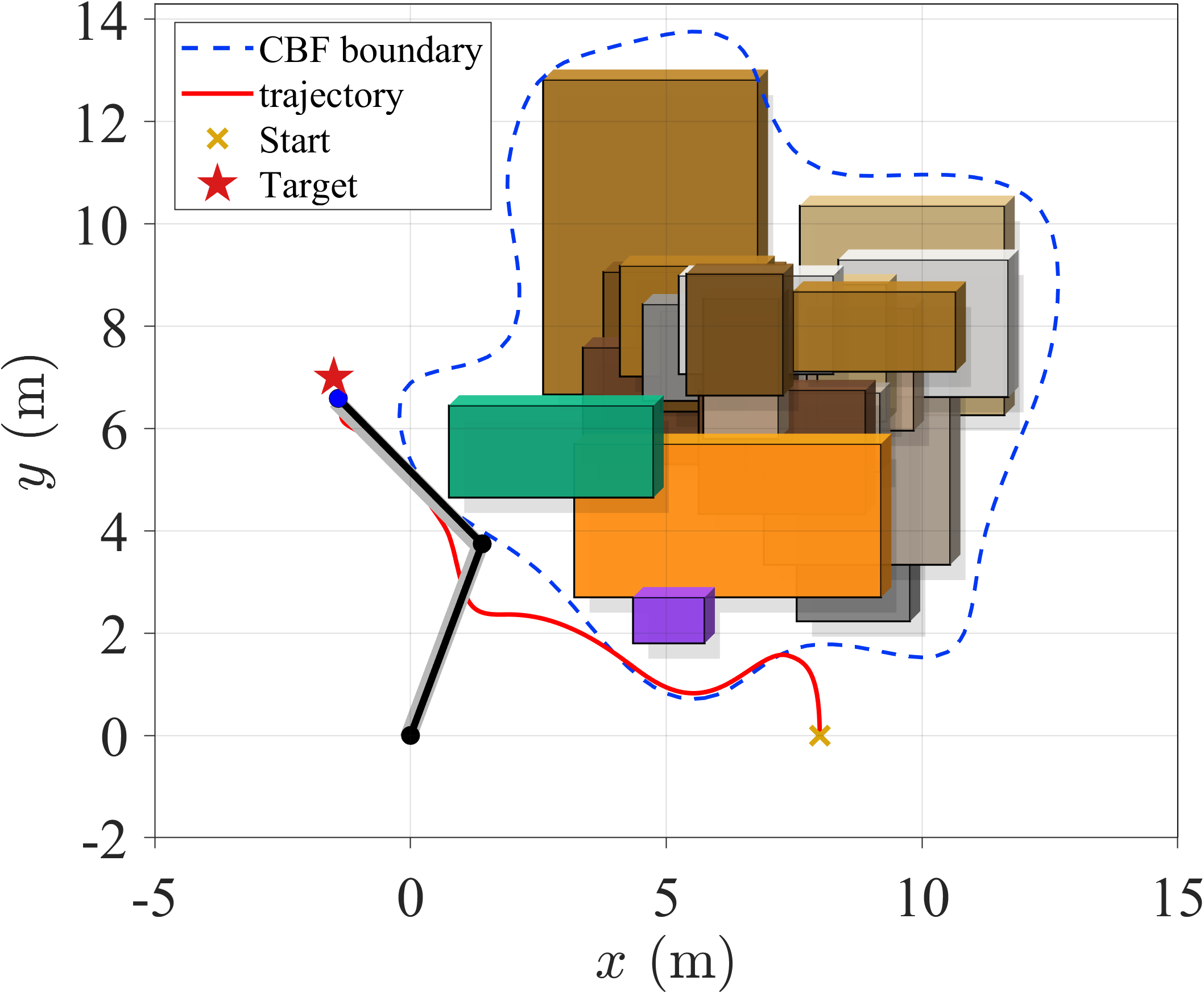}
        \par
        (i) $t=19.00$ s
    \end{minipage}

    \caption{
Endpoint trajectories during the SafeLink adaptation process:
(a) initial state;
(b)--(d), (e)--(f), and (g)--(h) show the first, second, and third unsafe-region expansions and subsequent adaptations, respectively; and
(i) final state upon reaching the target.
}
    \label{fig:trajectory}
\end{figure*}
\subsection{Results}
Model predictive control (MPC) is employed to generate reference control inputs, with the cost function defined as
\begin{equation}
\begin{aligned}
J = \sum_{k=1}^T \left\| \bm{p}_k - \bm{p}_{\text{target}} \right\|^2
+ \lambda_u \sum_{k=1}^T \|\bm{u}_k\|^2
+ \lambda_T \left\| \bm{p}_{T+1} - \bm{p}_{\text{target}} \right\|^2,
\end{aligned}
\end{equation}
where $T = 20$ denotes the prediction horizon, $\lambda_u>0$ and $\lambda_T>0$ are the control-effort and terminal-state weighting coefficients, respectively, and $\bm{p}_k$ and $\bm{p}_{\text{target}}$ represent the predicted and target endpoint positions, respectively.

To further enhance safety and smoothness, the angular velocities $\omega_1$ and $\omega_2$ are constrained by $|\omega_i|\leq \omega_{\max}$, $i=1,2$, using additional CBF constraints, where $\omega_{\max}=0.3$ rad/s. The resulting control constraints are expressed as
\begin{equation}
\label{eq:constraints_all}
\left[
\begin{array}{*4r}
0 & 0 & 1 & 0 \\
0 & 0 & -1 & 0 \\
0 & 0 & 0 & 1 \\
0 & 0 & 0 & -1
\end{array}
\right]\bm{g}(\bm{z})\bm{u}\leq
\begin{bmatrix}
\omega_{\max}-\omega_1\\
\omega_{\max}+\omega_1\\
\omega_{\max}-\omega_2\\
\omega_{\max}+\omega_2
\end{bmatrix}.
\end{equation}

\begin{figure}[htbp]
    \centering
    \begin{center}
\includegraphics[width=\linewidth]{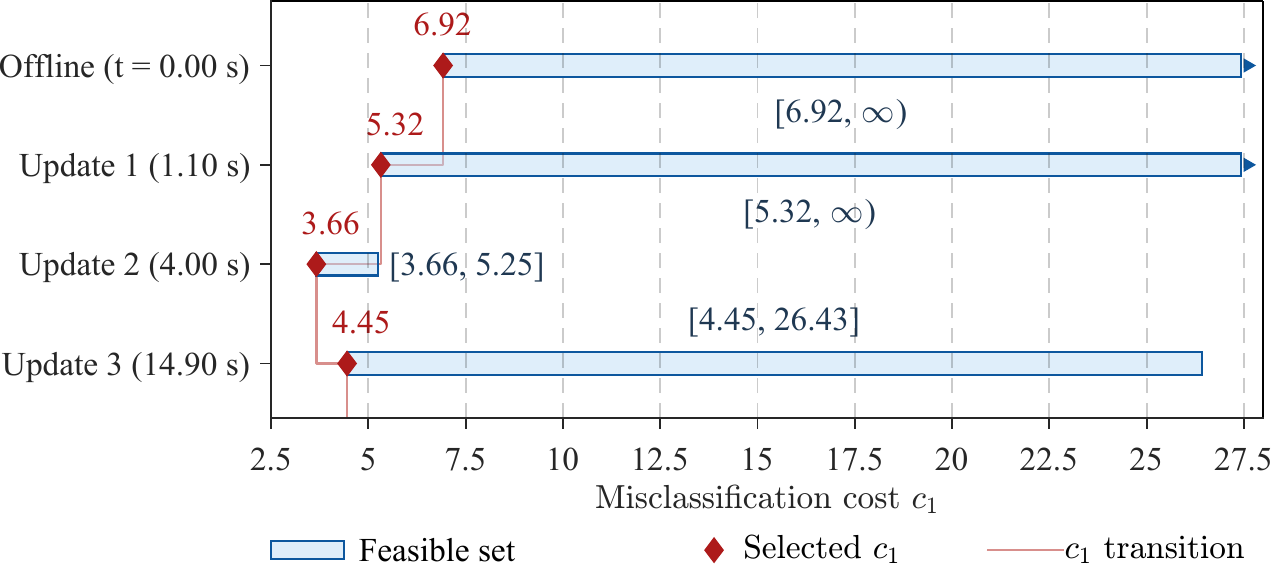}
    \end{center}
\caption{Compatible cost sets of the misclassification cost $c_1$ at the offline stage and after each update, together with the selected values and their stepwise evolution.}
    \label{fig:c1_intervals}
\end{figure}

\begin{figure}[htbp]
    \centering
    \begin{center}
\includegraphics[width=0.4\textwidth]{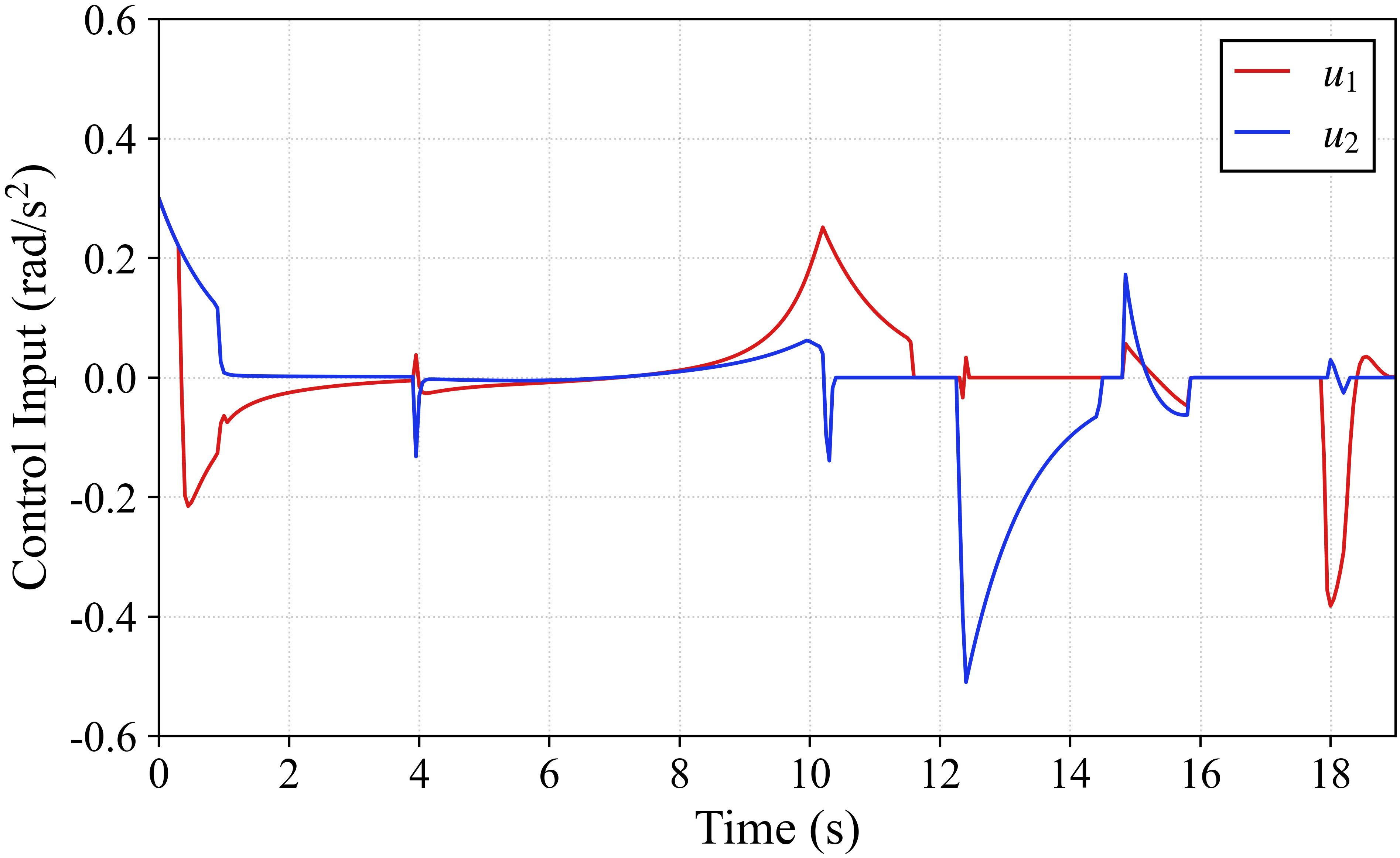}  
    \end{center}
\caption{Time evolution of the safe control inputs $u_1$ and $u_2$.}
    \label{fig:control_input}
\end{figure}

The safe control input \(\bm{u}_{safe}\) is obtained by solving the QP in Eq.~(\ref{eq:solve}) subject to the constraints in Eq.~(\ref{eq:constraints_all}).  As shown in Fig.~\ref{fig:control_input}, the control inputs remain within the prescribed bounds of $[-2,2]$ rad/s$^2$ throughout the simulation. The endpoint trajectory is illustrated in Fig.~\ref{fig:trajectory}. As shown in Fig.~\ref{fig:trajectory}, the learned CBF rapidly adapts after each sudden unsafe-region expansion and updates its boundary to cover the newly added unsafe samples, while the endpoint trajectory remains collision-free. Fig.~\ref{fig:c1_intervals} shows the compatible cost sets at the offline stage and after each update, together with the selected values of $c_1$. The selected cost is adjusted from $6.92$ to $5.32$, $3.66$, and $4.45$ after the three updates, respectively, so that it remains within the corresponding compatible cost set.

\begin{figure}[htbp]
    \begin{center}
\includegraphics[width=0.35\textwidth]{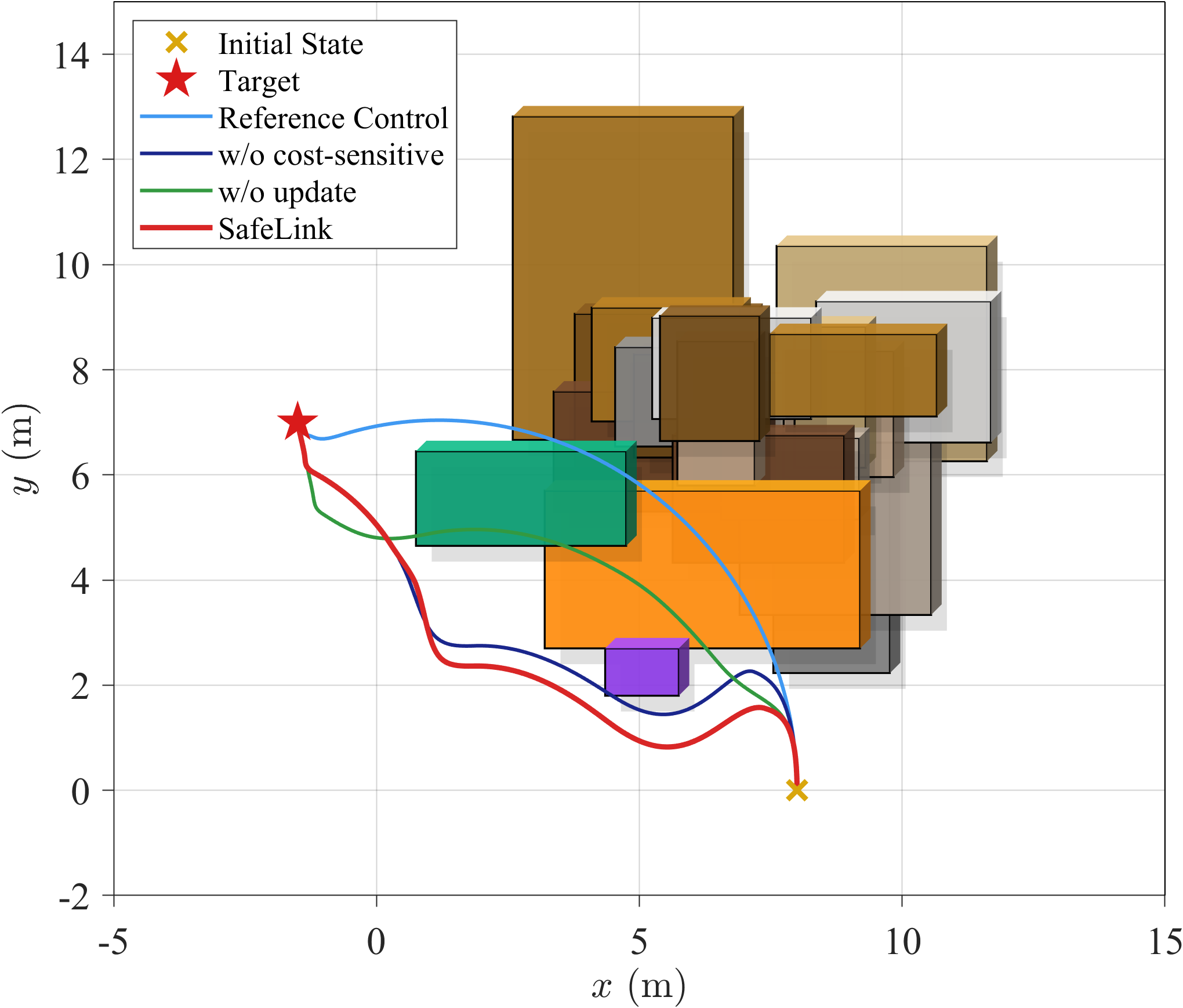}  
    \end{center}
\caption{Trajectories of the endpoint, where ``w/o'' denotes the absence of a specific part.}
    \label{fig:compared}
\end{figure}

For comparison, Fig.~\ref{fig:compared} presents the system trajectories under three scenarios: the reference method (MPC without obstacle consideration), SafeLink without CBF updates, and SafeLink without the cost-sensitive method. Specifically, the reference MPC approach focuses solely on path optimality, leading it to directly traverse unsafe regions. SafeLink without CBF updates remains collision-free before the first unsafe-region expansion; however, the unchanged CBF fails to account for subsequent changes in the unsafe region, resulting in collisions during the trajectory. For SafeLink without cost-sensitive learning, the learned unsafe set fails to conservatively cover the real unsafe region. 
Consequently, the endpoint can remain outside the learned unsafe set while still entering the real unsafe region, resulting in collisions. 
In contrast, the proposed method consistently avoids collisions in all cases, demonstrating its effectiveness under evolving unsafe regions.

\subsection{Update Efficiency Comparison}
Finally, the training and update runtimes of SafeLink are compared with those of an SVM-based CBF \cite{srinivasan2020synthesis} and an MLP-based CBF \cite{robey2020learning}. For a fair comparison, the three methods are configured to achieve comparable classification accuracies of approximately $92\%$ on the offline training set. The SVM uses a Gaussian kernel with a kernel scale of 10 and a box constraint of 3. The MLP consists of three hidden layers with 30 neurons each and is trained for 300 epochs with a learning rate of 0.01. SafeLink is configured with $M=100$, $\lambda=0.001$, $c_1=10$, and $c_2=1$. Here, $c_1=10$ is used to obtain a classification accuracy comparable to those of the SVM- and MLP-based CBFs. All methods use the same 5,792-sample offline training set and undergo the same three online updates as in the main experiment. The SVM and MLP are retrained from scratch after each update, whereas SafeLink performs the analytical decremental and incremental updates derived in Section~\ref{subsec:safelink_update}. All runtime measurements are performed in MATLAB on a platform with an Intel i5-13600KF CPU (14 cores, 3.50 GHz, 20 threads) and 32 GB of RAM.

\begin{figure}[htbp]
    \begin{center}
\includegraphics[width=0.45\textwidth]{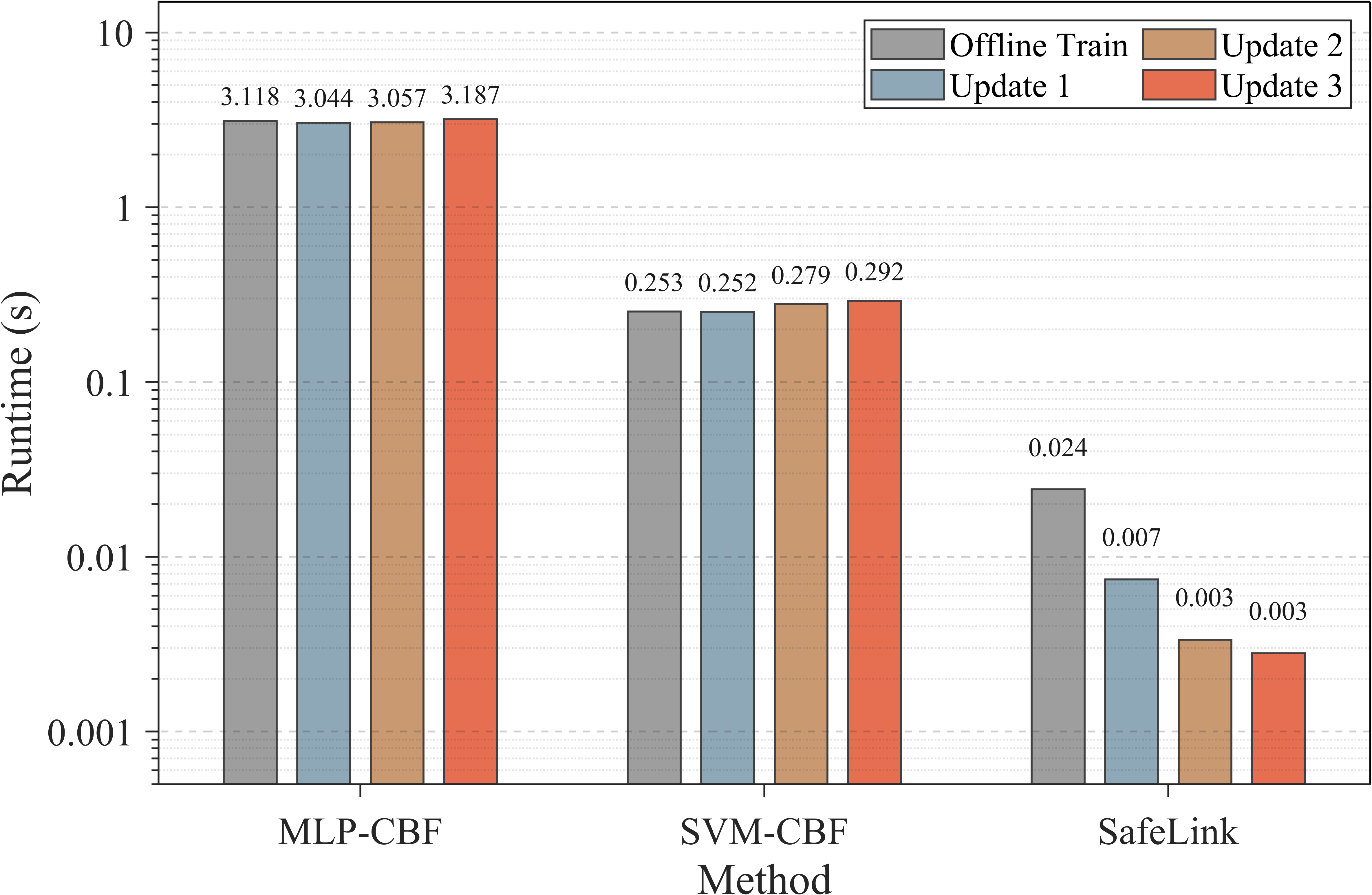}  
    \end{center}
\caption{Training and update runtimes of MLP-CBF, SVM-CBF, and SafeLink during offline training and the three online updates (log scale).}
    \label{fig:update_time}
\end{figure}

As shown in Fig.~\ref{fig:update_time} and Table~\ref{tab:time}, SafeLink achieves the lowest runtime at every stage. In particular, its online updates are approximately one to two orders of magnitude faster than those of SVM-CBF and about three orders of magnitude faster than those of MLP-CBF. Moreover, the analytical updates of SafeLink depend primarily on the numbers of changed samples rather than requiring retraining over the entire accumulated dataset. This leads to substantially lower online update costs when only a small fraction of samples changes at each update, making SafeLink suitable for rapid online CBF adaptation in safety-critical control applications.

\begin{table}[htbp]
    \centering
    \caption{Training and update performance over five runs. Runtimes (s), accuracies and recalls are reported as mean $\pm$ std.}
    \label{tab:time}
    \resizebox{\linewidth}{!}{
    \begin{tabular}{@{}l c c c@{}}
        \toprule
        \diagbox{Metric}{Method} & MLP-CBF & SVM-CBF & SafeLink\\
        \midrule
        Offline Training & 3.1179 $\pm$ 0.0790 & 0.2531 $\pm$ 0.0159 & \textbf{0.0243 $\pm$ 0.0110}\\
        Update 1         & 3.0438 $\pm$ 0.0877 & 0.2524 $\pm$ 0.0124 & \textbf{0.0074 $\pm$ 0.0095}\\
        Update 2         & 3.0570 $\pm$ 0.0763 & 0.2792 $\pm$ 0.0016 & \textbf{0.0034 $\pm$ 0.0029}\\
        Update 3         & 3.1866 $\pm$ 0.1095 & 0.2919 $\pm$ 0.0357 & \textbf{0.0028 $\pm$ 0.0026}\\
        Accuracy         & 0.9200 $\pm$ 0.0070 & 0.9223 $\pm$ 0.0000 & 0.9217 $\pm$ 0.0046\\
        Recall           & 1.0000 $\pm$ 0.0000 & 1.0000 $\pm$ 0.0000 & 1.0000 $\pm$ 0.0000\\
        \bottomrule
    \end{tabular}}
\end{table}

\section{Conclusion}
\label{sec:con}
In this paper, we developed SafeLink, a data-driven safety-critical control framework for evolving unsafe regions. SafeLink combines cost-sensitive RVFL learning with analytical adaptation to construct differentiable CBFs and efficiently update them as the unsafe region changes. We established sufficient conditions for conservative unsafe-region coverage and interval-wise safety, and experiments on a two-link manipulator demonstrated collision-free adaptation with substantially lower update runtimes. Future work will investigate full-horizon safety guarantees during unsafe-region evolution and adaptation to multiple disconnected unsafe regions.

\bibliographystyle{ieeetr}
\bibliography{References}

\end{document}